\documentclass{article}
\usepackage{iclr2027_conference,times}

\usepackage{amsmath,amssymb,amsthm}
\usepackage{graphicx}
\usepackage{xcolor}
\usepackage{hyperref}
\usepackage{booktabs}
\usepackage{caption}
\usepackage{subcaption}
\usepackage{algorithm}
\usepackage{algpseudocode}
\usepackage{comment}
\usepackage{tikz}
\usepackage{enumitem}
\usepackage{wrapfig}
\setlist[itemize,enumerate]{leftmargin=*, topsep=3pt, itemsep=1pt, parsep=0pt}
\usetikzlibrary{arrows.meta,shapes.geometric,positioning,calc,decorations.pathreplacing}

\hypersetup{colorlinks=true,linkcolor=blue!60!black,citecolor=green!50!black,urlcolor=blue}

\newtheorem{proposition}{Proposition}

\title{Think Short, Defer Smart, Act, and Repeat: Calibrated Reasoning and Uncertainty-Aware Deferral for Edge LLM Agents}

\author{
Amirmohammad Farzaneh \qquad Osvaldo Simeone \\
Institute for Intelligent Networked Systems (INSI) \\
Northeastern University London \\
London, UK \\
\texttt{\{a.farzaneh,o.simeone\}@nulondon.ac.uk}
}

\iclrfinalcopy
\begin{document}
\maketitle
\lhead{}

\begin{abstract}
LLM agents following the ReAct paradigm \citep{yao2022react} are promising enablers of complex multi-step tasks, including multi-hop question answering, code generation, and control of physical AI systems. Yet, when deployed at the edge, they must tightly manage their reasoning budget while remaining reliable and deferring to a cloud-side model only when local uncertainty is too high to act safely.
We propose \textbf{Think Short, Defer Smart (TSDS)}, a framework that synergistically
integrates a lightweight convergence probe, which halts on-device reasoning once the
intended action has stabilized, with a perplexity-based deferral rule that escalates
uncertain actions to a cloud-side model.
Both mechanisms are jointly calibrated on end-to-end episode trajectories via a
multi-objective Learn-Then-Test (LTT) procedure, providing simultaneous finite-sample
guarantees on expected episode reward and cloud-call rate.
We evaluate TSDS on four ReAct benchmarks spanning arithmetic reasoning (GSM8K), multi-hop
question answering (HotpotQA), code generation (MBPP), and multi-step embodied planning
(household robot), and compare against thought-calibration-only and calibrated-deferral-only
standalone baselines.
TSDS reduces per-episode thinking compute by $43\%$--$65\%$ over deferral-only baselines
across HotpotQA, MBPP, and the household robot task, while maintaining certified reward
and cloud-call rate guarantees.
\end{abstract}

\section{Introduction}
\label{sec:intro}

\begin{figure}[t]
\centering
\begin{subfigure}[b]{\textwidth}
  \centering
  \includegraphics[width=\textwidth]{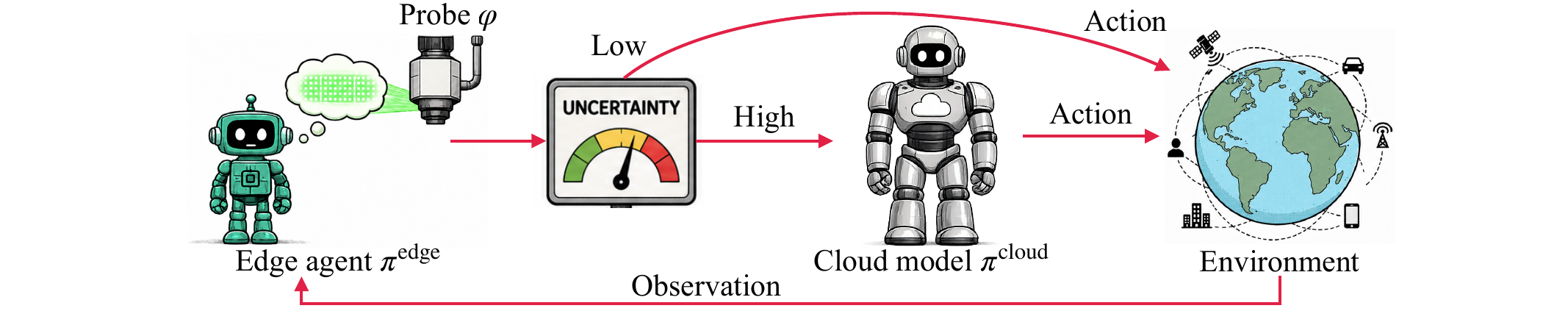}
  \caption{}
\end{subfigure}

\vspace{4pt}

\begin{subfigure}[b]{\textwidth}
  \centering
  \includegraphics[width=\textwidth]{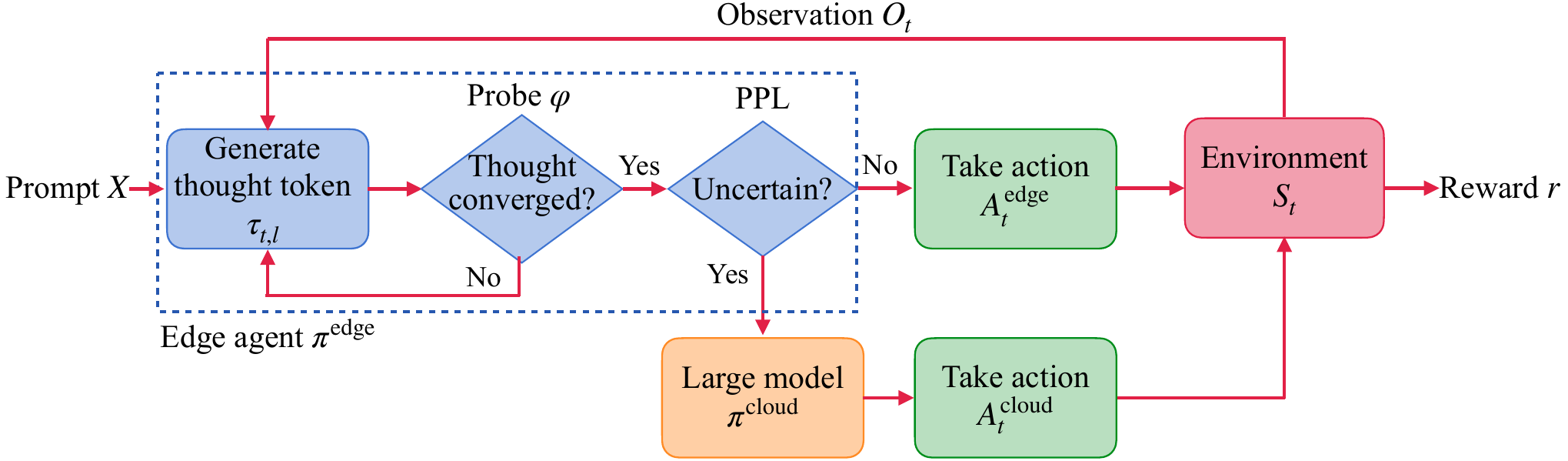}
  \caption{}
\end{subfigure}

\vspace{4pt}

\begin{subfigure}[b]{\textwidth}
  \centering
  \includegraphics[width=\textwidth]{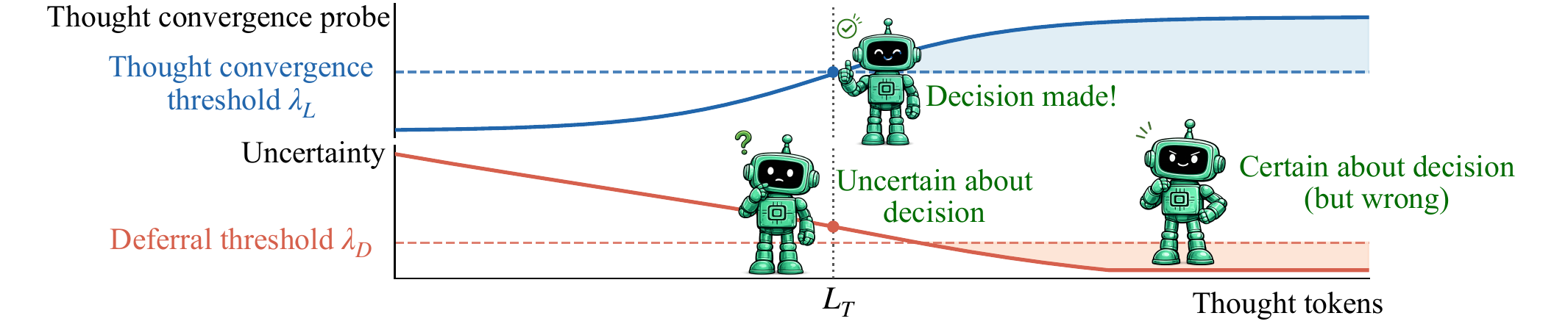}
  \caption{}
\end{subfigure}
\caption{TSDS pipeline as a high-level illustration (\textbf{a}) and detailed pipeline (\textbf{b}): An edge agent $\pi^\mathrm{edge}$ interacts with the environment across a number of time steps $t = 1, 2, \ldots$ At each step $t$, an action is taken after a reasoning phase, obtaining an observation from the environment. In the proposed TSDS, a convergence probe $\varphi$
monitors the hidden reasoning state and halts generation once the intended action has stabilized as gauged via a threshold $\lambda_L$. Then, an uncertainty measure on the chosen action, such as the PPL, is calculated: if it is higher than a threshold $\lambda_D$, the decision on the action is deferred to a larger cloud model $\pi^\mathrm{cloud}$. The chosen action, namely $A_t^\mathrm{edge}$ or $A_t^\mathrm{cloud}$, is executed in the environment, and an observation $O_t$ is collected. A final reward $r$ is produced at the end of an episode.
\textbf{(c)}~Illustration of the interplay between thought convergence and uncertainty-based deferral rule for a setting in which the agent selects a suboptimal action: The probe score $\varphi$ (top) crosses the thought convergence threshold $\lambda_L$ at $\tau^*$ tokens, showing that the edge agent's action has already converged. At this point, the uncertainty of the chosen action (bottom) is larger than the deferral threshold $\lambda_D$, correctly triggering deferral to the cloud agent. If reasoning were allowed to continue past token $\tau^*$, the edge agent would grow increasingly overconfident, suppressing the deferral mechanism and letting a suboptimal action through unchallenged (see Appendix \ref{app:interplay} for empirical evidence).}
\label{fig:framework}
\end{figure}

Autonomous agents powered by large language models (LLMs) are increasingly deployed in
resource-constrained settings, from robotic manipulation and autonomous navigation
\citep{ahn2022saycan,shridhar2021alfworld} to edge network management and industrial process
control \citep{xu2024ondevice,chevalier2023minigrid}.
In these settings, agents must operate on local hardware with limited compute, memory, and
power, while remaining reliable.
ReAct \citep{yao2022react} agents, which alternate free-text \emph{thoughts} with
executable \emph{actions}, serve as the backbone of such agentic systems
\citep{shinn2023reflexion,zhou2024language,sun2023adaplan}.
Thought generation consumes the majority of inference compute per step~\citep{wu2025thought,guo2025deepseek}.
However, even a well-reasoned action from an edge-side model can be wrong, and errors compound over the episode, making deferral to a cloud-based larger model necessary \citep{piatrashyn2026redact}.
In settings where cloud queries carry latency and bandwidth costs, an agent must therefore maintain reliability, while minimizing wasted reasoning compute and cloud-based calls.

As outlined in Fig. \ref{fig:framework}, our work addresses this problem by building on the following observations. First, in reasoning models, thinking may continue well beyond the point where a decision on intended action has converged.  \citet{esakkiraja2026therefore} show that the
committed action is encoded by hidden representations well before the end of the reasoning trace,
implying that many generated tokens are pure post-convergence waste. To make matters worse, as exemplified in Fig. \ref{fig:framework}, excessive thinking may lead the model to become increasingly overconfident in the selection of a suboptimal action. This complicates cloud deferral mechanisms, which are often based on the edge model's internal uncertainty signals \citep{malinin2021uncertainty,chen2024frugalgpt,piatrashyn2026redact}. Overall, a joint calibration of reasoning and deferral mechanisms has the potential to reduce resource consumption in terms of number of edge tokens, while also enhancing the efficiency of cloud deferral.

\paragraph{Our contribution:}
We propose \textbf{Think Short, Defer Smart (TSDS)}, a novel reasoning-based agentic framework that integrates and jointly calibrates the two synergetic mechanisms illustrated in Fig. \ref{fig:framework}:

\begin{enumerate}
\item \textbf{Thought calibration:}  At each interaction step with the environment, following the ReAct framework \citep{yao2022react}, the edge agent reasons to determine an optimized next action. In order to avoid overthinking, a lightweight convergence probe monitors the hidden
representations of the edge agent's model during reasoning. The probe is trained to detect the convergence of the reasoning trace towards a given decision \citep{wu2025thought}.

\item \textbf{Uncertainty-Aware Deferral:}
Once reasoning has converged and the edge action has been extracted, an uncertainty score over the resulting action token distribution is computed, and deferral to a cloud agent is triggered selectively by comparing the uncertainty score against a calibrated threshold $\lambda_D$ \citep{piatrashyn2026redact,chen2024frugalgpt}.

\end{enumerate}

Deciding when to stop thinking and when to defer entails a trade-off between the agent's performance in the environment, which is measured by the task-specific reward signal $r$, and resource consumption, which encompasses number of thinking steps, deferral rate, and number of environment interactions.
Furthermore, as illustrated in Fig. \ref{fig:framework}(c), thought calibration and cloud deferral are strongly intertwined, as excessive reasoning can produce over-confident uncertainty signals, causing the edge agent to select suboptimal actions, instead of correctly deferring to the cloud agent. Conversely, early thought truncation may produce an uncertainty signal higher than warranted by the actual confidence of the edge agent, unduly increasing deferral frequency. Joint calibration of reasoning and cloud deferral is carried out via \emph{Learn-Then-Test} (LTT) \citep{angelopoulos2021learn,farzaneh2026statistically}, which formulates hyperparameter selection as a multiple hypothesis testing problem, providing distribution-free finite-sample guarantees.

\section{Related Work}
\label{sec:related}

\paragraph{LLM agents:}
ReAct \citep{yao2022react} established a by-now canonical pattern of alternating free-text reasoning
with tool calls. Subsequent work added verbal self-reflection \citep{shinn2023reflexion}, adaptive
planning \citep{sun2023adaplan}, and lifelong skill libraries \citep{wang2023voyager}.
A fundamental limitation shared by all these systems is that each thought trace runs until a
positional stop string, and hence inference cost scales with the trace length budget rather than with
the actual difficulty of the reasoning step.

\paragraph{Uncertainty quantification and deferral:}
Information-theoretic scores such as perplexity (PPL) provide uncertainty estimates that correlate with actual model
errors \citep{malinin2021uncertainty,fomicheva2020unsupervised}. However, translating these raw scores into reliable deployment-level guarantees
requires explicit calibration \citep{angelopoulos2021learn}.
Model cascades \citep{chen2024frugalgpt,yue2023llm,ong2025routellm, hou2026reliable} route single-turn queries
across models of different capacities, focusing on independent requests without multi-step
dependencies.
ReDAct \citep{piatrashyn2026redact} applies PPL-guided deferral inside an
agentic loop, but the deferral threshold is chosen heuristically.

\paragraph{Test-time compute and thought calibration:}
Thought calibration \citep{wu2025thought} trains a convergence probe on offline single-turn
trajectories and uses LTT \citep{angelopoulos2021learn} to certify a stopping threshold.
Early-exit methods \citep{yang2026dynamic} and conformal prediction wrappers
\citep{chen2024knowing} also aim at an adaptive use of inference resources. Unlike TSDS, however, none of these methods jointly calibrate a stopping rule with a deferral mechanism under formal statistical guarantees, making these methods unsuitable as direct baselines in our setting.

\section{Problem Definition}
\label{sec:background}

\paragraph{Edge agent and environment:}
\label{sec:setup}
As shown in Fig.~\ref{fig:framework}, we consider an edge agent $\pi^\mathrm{edge}$ that interacts with a physical or virtual environment over discrete time steps $t =1, \ldots, T$ to carry out a task described by a natural language prompt $X$.
At each step $t$, the agent maintains a history
\[
  H_t = \bigl(X,\; (O_1, T_1, A_1),\; \ldots,\; (O_{t-1}, T_{t-1}, A_{t-1}),\; O_t\bigr),
\]
where $(T_t, A_t)$ are the thought trace and action at step~$t$, and $O_t$ is the observation of the environment upon taking action $A_{t-1}$.
Actions may include tool calls (e.g., $\mathtt{Search}$, $\mathtt{Lookup}$), code execution, free-form text, or motor commands in a physical environment~\citep{ahn2022saycan,shridhar2021alfworld,chevalier2023minigrid}.
At episode termination, i.e., after time $t = T$, the environment issues a scalar reward $r(X, A_1, \ldots, A_T) \in [0,1]$ that depends on the task prompt $X$, which specifies both the goal and the evaluation criterion, and on the trajectory of actions taken. Examples include exact-match for question answering~\citep{yang2018hotpotqa,cobbe2021gsm8k} and binary pass/fail for goal-oriented tasks~\citep{austin2021program,shridhar2021alfworld}.

At each time step $t$, given history $H_t$, the edge agent generates a thought trace $T_t = (\tau_{t,1}, \ldots, \tau_{t,L_t})$. Each token $\tau_{t,i} \sim \pi^\mathrm{edge}(\cdot \mid H_t, \tau_{t,1}, \ldots, \tau_{t,i-1})$ is drawn autoregressively using the edge agent's model $\pi^\mathrm{edge}$. For future reference, we denote as $h_{t,i}^{(\ell)}\in \mathbb{R}^d$ the latent activation of model $\pi^\mathrm{edge}$ for token $i$ at layer $\ell$. After $L_t$ reasoning steps, the edge agent may take an action $A_t^\mathrm{edge}$ or defer to the cloud. The action $A_t^\mathrm{edge}$ is produced by model $\pi^\mathrm{edge}$ as $A_t^\mathrm{edge} \sim \pi^\mathrm{edge}(\cdot\mid  H_t, \tau_{t,1}, \ldots, \tau_{t,i-1})$.

\paragraph{Cloud agent and cascading:}
\label{sec:redact}
A more capable cloud model $\pi^\mathrm{cloud}$ is available to the edge agent as a fallback. If a cloud deferral decision is made at time $t$, the full history $H_t$ is transferred to the cloud, and the model
$\pi^\mathrm{cloud}$ generates an alternative action
$A_t^\mathrm{cloud} \sim \pi^\mathrm{cloud}(\cdot \mid H_t)$,
replacing the edge agent's action $A_t^\mathrm{edge}$ for that step.
We write as $D_t \in \{0,1\}$ the indicator for whether the cloud model is called ($D_t = 1$) or not ($D_t = 0$).

\paragraph{Thinking convergence and deferral:}
The edge agent's operation is governed by two threshold-parameterized mechanisms.
A \emph{thinking convergence rule} decides at each thought position $i$ whether to halt reasoning and extract an action $A_t^\mathrm{edge}$ and an uncertainty score $U_t$. To this end, the rule applies a threshold $\lambda_L$ to a convergence signal $C_{i,t}$ extracted from the latent activations $h_{t,i}^{(\ell)}$ of the edge model as: stop if $C_{t,i} > \lambda_L$ and continue otherwise.
A \emph{deferral rule} decides whether to execute the edge action or hand the step to $\pi^\mathrm{cloud}$ using a threshold $\lambda_D$ as: defer if $U_t >\lambda_D$ and not defer otherwise. Accordingly, the executed action $A_t$ is
\begin{equation}
    A_t =
    \begin{cases}
        A_t^\mathrm{edge}\quad \text{if} \; U_t\leq \lambda_D\\
        A_t^\mathrm{cloud}\quad \text{if} \; U_t > \lambda_D,
    \end{cases}
    \label{eq:action}
\end{equation}
with action $A_t^\mathrm{edge}$ produced at thinking step
\begin{equation}
    L_t = \min\{\arg\min_i \{i:C_{t,i} > \lambda_L\}, L_\mathrm{max}\},
    \label{eq:stopping}
\end{equation}
where $L_\mathrm{max}$ is the maximum allowed number of thinking steps.

\paragraph{Design objectives:}
We define the expected episode reward
$R(\boldsymbol{\lambda}) = \mathbb{E}[r(\boldsymbol{\lambda})]$,
where the expectation is over task prompts and environment stochasticity, and we have highlighted the dependence on the vector $\boldsymbol{\lambda} = (\lambda_L, \lambda_D)$ of hyperparameters.
The \emph{deferral rate}, i.e., the expected fraction of agent steps
deferred to the cloud, and the thinking cost, i.e., number of reasoning tokens per episode, are given by
\begin{equation}
C_D(\boldsymbol{\lambda})
=
\mathbb{E}\!\left[
\frac{1}{T}\sum_{t=1}^{T}D_t
\right] \;\; \text{and}\;\;C_L(\boldsymbol{\lambda}) = \mathbb{E}\!\left[\sum_{t=1}^T L_t\right],
\label{eq:deferral_cost_bg}
\end{equation}
respectively.
Another important metric is the \emph{episode-length cost}
$C_S(\boldsymbol{\lambda}) = \mathbb{E}[T]$,
where $T$ is the (random) number of agent steps until episode termination.
The overall design objective is to minimize thinking cost $C_L(\boldsymbol{\lambda})$, while keeping average reward $R(\boldsymbol{\lambda})$ above a floor $R^{\min}$ and cloud deferral rate $C_D(\boldsymbol{\lambda})$ within a budget $C_D^{\max}$, i.e.,
\begin{equation}
\min_{\boldsymbol{\lambda}}\; C_L(\boldsymbol{\lambda})
\quad \text{subject to} \quad
R(\boldsymbol{\lambda}) \geq R^{\min} \;\;\text{and}\;\;C_D(\boldsymbol{\lambda}) \leq C_D^{\max}.
\label{eq:objective}
\end{equation}

\section{Think Short, Defer Smart}
\label{sec:tc}

This section introduces TSDS, a framework that addresses problem~\eqref{eq:objective} by proposing concrete instantiations of the thinking convergence rule (Sec.~\ref{sec:probe}) and the deferral rule (Sec.~\ref{sec:deferral_mechanism}), as well as by incorporating the calibration of their joint threshold vector $\boldsymbol{\lambda}$ via multi-objective LTT (Sec.~\ref{sec:multi_obj_calibration}).

\subsection{Convergence Probe and Thought Calibration}
\label{sec:probe}

The goal of the thinking convergence rule is to detect, from the edge agent $\pi^\mathrm{edge}$'s internal representations, the moment at which its reasoning has committed to a stable action.
We follow the general approach of \citep{wu2025thought} of training a lightweight probe on offline trajectories to certify a stopping threshold, tailoring the methodology to agents that \emph{reason to act} in a multi-step interactive loop.

As illustrated in Fig.~\ref{fig:framework}, define the convergence signal as $C_{t,i} = \varphi_\theta(h_{t,i}^{(\ell)}) \in \mathbb{R}$, thus mapping the hidden state $h_{t,i}^{(\ell)}$ at reasoning step $i$ and layer $\ell$ to the scalar convergence score $C_{t,i}$.
The layer $\ell$ is treated as a hyperparameter. The probe $\varphi_\theta$ is trained offline on a dataset $\mathcal{D}_{\mathrm{tr}} = \{(h_{t,i}^{(\ell),(n)},\, y_{t,i}^{(n)})\}_{n = 1}^{N_\mathrm{tr}}$, where index $n = 1,\ldots,N_{\mathrm{tr}}$ runs over training episodes. The dataset $\mathcal{D}_{\mathrm{tr}}$ is collected by running the model $\pi^\mathrm{edge}$ with \emph{full} thought generation (no early stopping, no deferral) and by recording for each position $i$ in a set $\mathcal{I}$ the latent state $h_{t,i}^{(\ell)}$. Position $i$ is labelled positive, i.e., $y_{t,i}^{(n)} = 1$, if the early-stopped action at every subsequent probe position $i' \geq i$ within the same step agrees with the full-thought action, i.e.,
\begin{equation}
y_{t,i}^{(n)} \;=\; \prod_{i' \in \mathcal{I}: i' \geq i} \mathbf{1}\bigl[A_{i',t}^{(n)} = A_t^{(n)}\bigr] \in \{0, 1\},
\label{eq:probe_label}
\end{equation}
where $A_t^{(n)}$ is the full-thought reference action and $A_{i,t}^{(n)}$ is the early-stopped action obtained by truncating the trace at position $i$ and injecting token \texttt{</think>}.

\subsection{Cloud Deferral}
\label{sec:deferral_mechanism}

Once the edge action $A_t^\mathrm{edge}$ is extracted, the uncertainty score $U_t$ is obtained as the PPL of the edge model for action $A_t^\mathrm{edge}$, following \citep{piatrashyn2026redact}. Alternative scores are defined in Appendix~\ref{app:uncertainty} and can be substituted without modifying the calibration procedure.

\subsection{Multi-Objective Calibration}
\label{sec:multi_obj_calibration}

In this section, we address the multi-objective optimization problem~\eqref{eq:objective} via LTT \citep{angelopoulos2021learn,farzaneh2026statistically}.
Following LTT, we construct a finite grid $\Lambda = \{(\lambda_L^{(i)}, \lambda_D^{(j)})\}$ of $|\Lambda|$ candidate pairs.
For each candidate pair $\lambda=(\lambda_L,\lambda_D)\in\Lambda$, we
run TSDS with both thought calibration and deferral active and obtain
the held-out calibration dataset
\begin{equation}
\mathcal{D}_{\mathrm{cal}}(\lambda)
=
\left\{
\left(r^{(n)}(\lambda),c^{(n)}(\lambda)\right)
\right\}_{n=1}^{N_{\mathrm{cal}}},
\label{eq:calibration_dataset}
\end{equation}
where $r^{(n)}(\lambda)$ is the terminal reward of episode $n$ and
\begin{equation}
c^{(n)}(\lambda)
=
\frac{1}{T^{(n)}(\lambda)}
\sum_{t=1}^{T^{(n)}(\lambda)}
\mathbf{1}[U_t^{(n)}(\lambda)>\lambda_D],
\label{eq:episode_deferral_fraction}
\end{equation}
where $T^{(n)}(\lambda)$ denotes the number of agent steps taken before
termination of episode $n$ under policy parameters $\lambda$.
Thus, each calibration episode contributes one reward observation and
one deferral-rate observation, both bounded in $[0,1]$.

\begin{proposition}[Joint risk control guarantee \citep{angelopoulos2021learn}]
\label{prop:joint}
Assume that calibration data $\mathcal{D}_\mathrm{cal}$ is drawn i.i.d. from the same distribution underlying the generation of the test episode. Then, the hyperparameter $\boldsymbol{\hat{\mathbf{\lambda}}}$ returned by LTT satisfies the constraints in problem \eqref{eq:objective} with probability larger than $1-\delta$, i.e.,
\begin{equation}
P\!\bigl[R(\boldsymbol{\hat{\mathbf{\lambda}}}) \geq R^{\min} \;\text{ and }\; C_D(\boldsymbol{\hat{\mathbf{\lambda}}}) \leq C_D^{\max}\bigr]
\geq 1-\delta,
\label{eq:joint_guarantee}
\end{equation}
where $\delta$ is a user-defined probability.
\end{proposition}

When the certified set $\hat{\Lambda}$ is empty (no candidate pair clears both statistical tests), the agent falls back to E-ReAct: full thought traces with no deferral, $\lambda_L = \infty$. This conservative default never violates $C_D^{\max}$ and requires no calibrated parameter.

\begin{algorithm}[t]
\caption{Think Short, Defer Smart (step $t$)}
\label{alg:main}
\begin{algorithmic}[1]
\Require Certified pair $\boldsymbol{\lambda} = (\lambda_L, \lambda_D)
         \in \hat{\Lambda}$; probe $\varphi_\theta$; $\pi^\mathrm{edge}$;
         $\pi^\mathrm{cloud}$; history $H_t$.
\State Generate first token $\tau_{t,1}$ with $\pi^\mathrm{edge}$; $i \gets 1$
\While{$i < L_{\max}$}
    \State Retrieve hidden state $h_{t,i}$
    \If{$\varphi_\theta(h_{t,i}) \geq \lambda_L$} \textbf{break} \Comment{Thought converged}
    \Else{} generate next token $\tau_{t,i+1}$; $i \gets i + 1$
    \EndIf
\EndWhile
\State Inject \texttt{</think>}; generate $A_t^\mathrm{edge}$ \Comment{Action extraction}
\State $u_t \gets U(A_t^\mathrm{edge} \mid H_t, \tau_{t,1:i}, \texttt{</think>})$
\If{$u_t \leq \lambda_D$}
    \State $A_t \gets A_t^\mathrm{edge}$ \Comment{Confident: accept}
\Else
    \State $A_t^\mathrm{cloud} \sim \pi^\mathrm{cloud}(\cdot \mid H_t)$ \Comment{Uncertain: defer}
    \State $A_t \gets A_t^\mathrm{cloud}$
\EndIf
\State Execute $A_t$; observe $O_{t+1}$; update $H_{t+1}$
\end{algorithmic}
\end{algorithm}

\section{Experiments}
\label{sec:experiments}

We evaluate TSDS on three benchmarks: code generation (MBPP, Sec.~\ref{sec:exp_mbpp}), multi-step question answering (HotpotQA, Sec.~\ref{sec:exp_hotpotqa}), and simulated household-robot planning (Sec.~\ref{sec:exp_household}). An additional experiment on single-step arithmetic reasoning (GSM8K) can be found in Appendix~\ref{app:gsm8k_details}. The edge model \texttt{DeepSeek-R1-Distill-Qwen-7B} and the Exponential Moving Average (EMA) convergence probe are adopted throughout.
All experiments implement LTT using exact one-sided binomial p-values
for binary episode rewards and one-sided Hoeffding p-values for the
bounded episode-level deferral fractions, with Bonferroni correction at
level $\delta=0.10$ (see Appendix \ref{app:ltt}). The target deferral rate is $C_D^{\max}=0.70$.
The candidate grid $\Lambda$ for LTT is formed by considering a uniform sweep of the unit interval with $5$ levels for threshold $\lambda_L$ and $4$ uniformly spaced quantiles of the per-episode maximum uncertainty on dataset $\mathcal{D}_{\mathrm{tr}}$ for threshold $\lambda_D$.
Benchmark-specific settings and parameters are given in each subsection.
For computational efficiency, Algorithm~\ref{alg:main} is implemented with strided probe evaluation so that the convergence probe is queried every $s{=}16$ thought tokens rather than at every token.
All reported metrics are means over $50$ independent calibration--test splits.

\subsection{Baselines}
\label{sec:baselines}

We compare TSDS against the following policies, which isolate the contribution of different components.

\begin{itemize}
\item \emph{E-ReAct} (Edge ReAct): As in the original ReAct baseline \citep{yao2022react}, the model $\pi^\mathrm{edge}$ runs with full thought traces ($\lambda_L = \infty$), and there is no deferral ($\lambda_D = \infty$).

\item \emph{Cloud ReAct}: The cloud model $\pi^\mathrm{cloud}$ is used for every step, providing a performance ceiling at maximum cloud deferral cost.

\item \emph{E-ReAct-TC} (Edge ReAct with Thought Calibration): Thought calibration is active, but deferral is disabled ($\lambda_D = +\infty$).
  This isolates the contribution of thought truncation alone, without the safety net of the deferral mechanism.

\item \emph{ReDAct} \citep{piatrashyn2026redact}: The edge model $\pi^\mathrm{edge}$ generates full thought traces ($\lambda_L = \infty$), deferring whenever the PPL exceeds a fixed threshold $\hat{\lambda}_D$, set to the median PPL on $\mathcal{D}_{\mathrm{tr}}$.

\item \emph{ReDAct-CD} (ReDAct with Calibrated Deferral): The edge model $\pi^\mathrm{edge}$ generates full thought traces ($\lambda_L = \infty$), setting $L_t = L_{\max}$ in \eqref{eq:stopping}, and the deferral threshold $\lambda_D$ is selected via single-objective LTT on dataset $\mathcal{D}_{\mathrm{cal}}$ to satisfy the reward constraint $R(\boldsymbol{\lambda}) \geq R^{\min}$ with probability $1{-}\delta$.
  This isolates the contribution of calibrated deferral without thought calibration.
\end{itemize}

\subsection{Code Generation: MBPP}
\label{sec:exp_mbpp}

We evaluate TSDS on MBPP \citep{austin2021program}, a benchmark of mostly basic Python programming problems.
Each problem specifies a Python function in natural language with unit tests, and reward is $1$ if all unit tests pass and $0$ otherwise.
Actions are complete Python functions, with a mean of $1166$ thought tokens per problem.
This is a single-step setting, i.e., $T=1$.

\paragraph{Setup:}
The cloud model is \texttt{DeepSeek-R1-Distill-Qwen-32B}, and the thought caps for both the edge and cloud models is set to $L_{\max}=2048$.
The probe $\varphi_\theta$ is trained at layer $\ell=5$ on $100$ MBPP train-split episodes.
We evaluate on $257$ test problems with $50$ random $60/40$ splits.
The reasoning-threshold grid is $\lambda_L \in \{0.1, 0.2, \ldots, 0.9\}$, and the deferral grid $\lambda_D \in \{0.084, 0.092, 0.106, 0.117, 0.141\}$ [nat] is drawn from PPL percentiles on $\mathcal{D}_{\mathrm{tr}}$, giving $|\Lambda|=45$ candidate pairs.
The minimum reward floor is $R^{\min}=0.69$.

\begin{figure}[t]
\centering
\includegraphics[width=\textwidth]{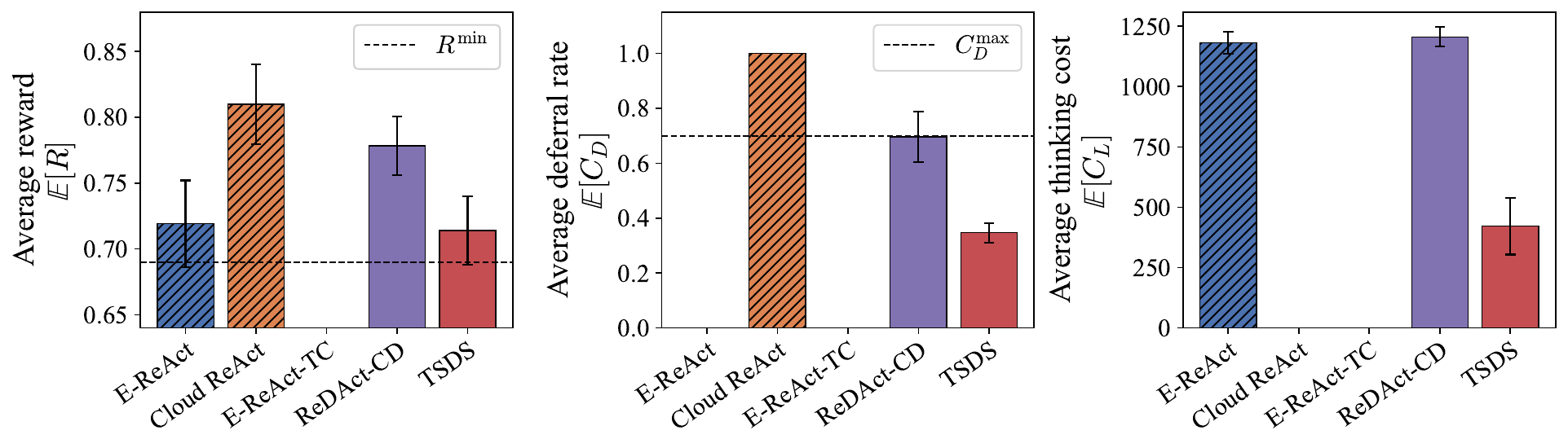}
\caption{Test results for MBPP, shown for five policies and averaged over $50$ random $60/40$ calibration--test splits ($n_{\mathrm{test}}=103$ per split). Error bars show one standard deviation across splits.
\emph{Left}: Average reward per policy. The dashed horizontal line marks the minimum reward floor $R^{\min}=0.69$.
\emph{Center}: Average deferral rate to the cloud model. The dashed line marks the maximum allowed deferral fraction $C_D^{\max}=0.70$.
\emph{Right}: Average thinking cost (tokens).
Hatched bars denote uncalibrated reference policies.}
\label{fig:exp4_test_bars}
\end{figure}

Fig.~\ref{fig:exp4_test_bars} reports average reward, deferral rate, and thinking cost for all policies, averaged over $50$ calibration--test splits.
E-ReAct runs full thought traces without deferral, establishing the uncalibrated edge baseline.
Cloud ReAct achieves higher reward but routes every problem to the cloud model, exhausting the deferral budget.
E-ReAct-TC is not able to certify any hyperparameters to guarantee reward above the floor $R^{\min}$ due to not having access to the cloud model.
ReDAct-CD improves reward through selective cloud escalation but exceeds the cloud-deferral budget on average and requires full thought traces for the edge agent.
TSDS clears the reward floor and respects the deferral budget while using $64\%$ fewer thought tokens than E-ReAct and ReDAct-CD and calling the cloud model on less than half the rate of ReDAct-CD.

\subsection{Multi-Step Setting: HotpotQA}
\label{sec:exp_hotpotqa}

We evaluate TSDS on HotpotQA \citep{yang2018hotpotqa}, a multi-hop QA benchmark where the
agent takes up to $T_{\max} = 7$ steps, with action space including the tools $\mathtt{Search}[\cdot]$, $\mathtt{Lookup}[\cdot]$,
and $\mathtt{Finish}[\cdot]$, to answer questions requiring two supporting passages.
Reward is a binary terminal exact match indicator.

\paragraph{Setup:}
The cloud model is \texttt{DeepSeek-R1-Distill-Qwen-14B}. Edge and cloud thought caps are set to $L_{\max}=384$ and $1024$, respectively.
The probe $\varphi_\theta$ is trained at layer $\ell=5$ on $50$ train-split episodes.
We evaluate on $250$ validation problems with $50$ random $75/25$ splits.
The reasoning-threshold grid is $\lambda_L \in \{0.1, 0.3, 0.5, 0.7, 0.9\}$, and the deferral grid $\lambda_D \in \{0.258, 0.318, 0.355, 0.392\}$ [nat], corresponds to percentiles $\{40, 60, 75, 85\}$ of the per-episode maximum uncertainty on dataset $\mathcal{D}_{\mathrm{tr}}$, giving $|\Lambda|=20$ candidate hyperparameters.
The minimum reward floor is $R^{\min}=0.22$.

\begin{wrapfigure}{r}{0.5\textwidth}
\centering
\includegraphics[width=0.48\textwidth]{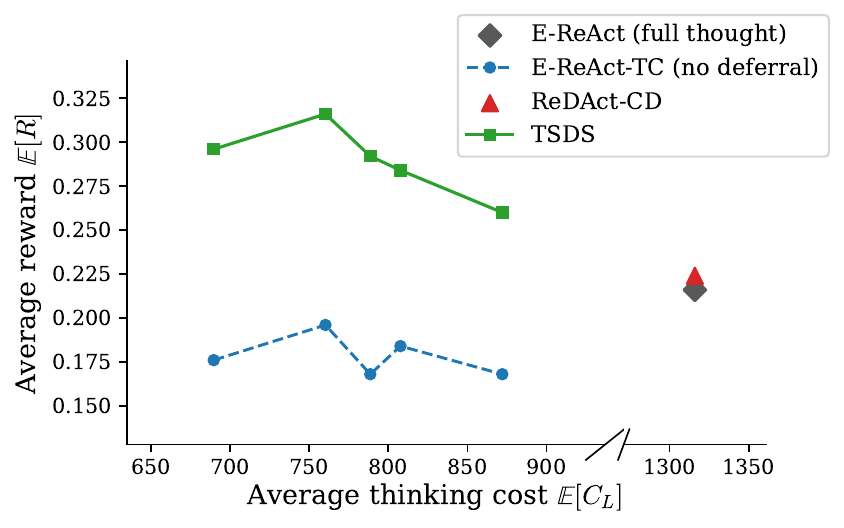}
\caption{Reward vs.\ thinking cost on HotpotQA across five $\lambda_L$ grid values.
The blue dashed curve shows thought calibration only, and the solid green curve shows TSDS at the certified deferral threshold $\lambda_D=0.318$.
All values are means over the full pool of $250$ validation episodes.}
\label{fig:exp3_tradeoff}
\vspace{-30pt}
\end{wrapfigure}

Fig.~\ref{fig:exp3_tradeoff} shows the average reward as a function of the average thinking cost, as obtained by varying the threshold $\lambda_L$ for a fixed deferral threshold $\lambda_D = 0.318$. Thought calibration is seen to significantly reduce the thinking cost relative to E-ReAct, but the average reward is negatively affected. In contrast, TSDS uniformly outperforms E-ReAct,
illustrating the benefit of combining thought calibration and cloud deferral. Additionally, ReDAct-CD only marginally improves over E-ReAct, confirming that untruncated reasoning induces overconfidence in the edge model, resulting in fewer actions being deferred to the cloud model.

\begin{figure}[b]
\centering
\includegraphics[width=\textwidth]{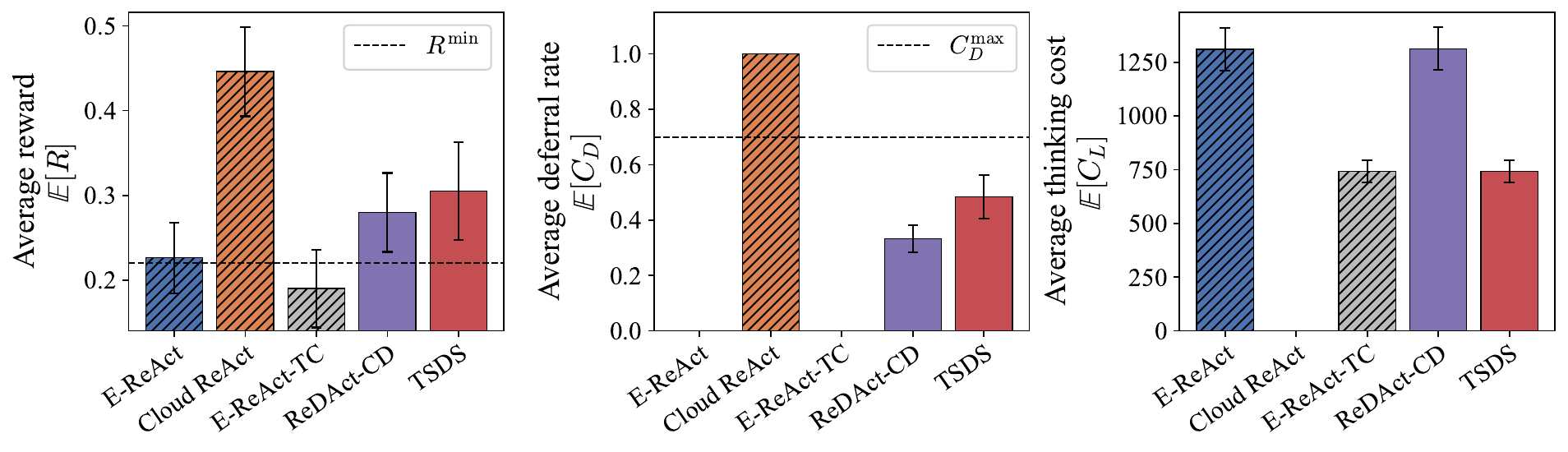}
\caption{Test results for HotpotQA, shown for all five policies and averaged over $50$ random $75/25$ calibration--test splits. Error bars show one standard deviation across splits.
\emph{Left}: Average reward per policy. The dashed horizontal line marks the minimum reward floor $R^{\min}=0.22$.
\emph{Center}: Average deferral rate to the cloud model. The dashed line marks the maximum allowed deferral fraction $C_D^{\max}=0.70$.
\emph{Right}: Average thinking cost (tokens).
Hatched bars denote uncalibrated reference policies.}
\label{fig:exp3_test_bars}
\end{figure}

Fig.~\ref{fig:exp3_test_bars} reports average reward, deferral rate, and thinking cost for all policies, averaged over $50$ calibration--test splits.
ReDAct-CD improves reward through selective cloud escalation, though at the cost of the full reasoning budget.
The pattern mirrors Fig.~\ref{fig:exp4_test_bars}. ReDAct-CD and TSDS are the only two policies clearing the reward floor and respecting the deferral budget, while TSDS does so with substantially fewer thought tokens than ReDAct-CD. In the multi-step setting of this experiment this advantage is further reinforced, as per-step thought calibration concentrates cloud deferral on precisely the steps where the edge model is most uncertain.

\subsection{Simulated Household-Robot Planning}
\label{sec:exp_household}

To evaluate TSDS in a physical AI setting, we introduce a text-based simulated
household-manipulation benchmark for evaluating multi-step physical-AI agents without
requiring vision or physical hardware.
Inspired by SayCan~\citep{ahn2022saycan} and ALFWorld~\citep{shridhar2021alfworld}, the
environment places an agent in a four-room setting (bedroom, kitchen, bathroom, living room)
and asks it to complete tasks drawn from four families, namely \emph{fetch}, \emph{place},
\emph{clean}, and \emph{heat}, by issuing five action primitives:
\texttt{go~to~<room>}, \texttt{pick~up~<object>}, \texttt{put~down~<object>},
\texttt{heat~<object>}, and \texttt{clean~<object>}, within at most $T_{\max}=6$ steps.
Reward is binary and is equal to $1$ if the task is completed by episode termination, and to $0$ otherwise.
The environment layout is illustrated in Appendix~\ref{app:household_details} (Fig.~\ref{fig:household_floorplan}).

\paragraph{Setup:}
The same 7B/14B model pair as HotpotQA is used with $L_{\max}=512$ for the edge agent and $1024$ for the cloud agent.
The probe is retrained at layer $\ell=10$ on $50$ household episodes (see Appendix~\ref{app:probe_layer_household} for an ablation study).
We evaluate on $200$ held-out episodes with $50$ random $75/25$ splits.
The reasoning-threshold grid is $\lambda_L \in \{0.1, 0.3, 0.5, 0.7, 0.9\}$, and the deferral grid $\lambda_D \in \{0.287, 0.313, 0.343, 0.384\}$ [nat] corresponds to percentiles $\{40, 55, 70, 85\}$ of the per-episode maximum uncertainty on $\mathcal{D}_{\mathrm{tr}}$, giving $|\Lambda|=20$.
The minimum reward floor is $R^{\min}=0.52$.

\begin{figure}[t]
\centering
\includegraphics[width=\textwidth]{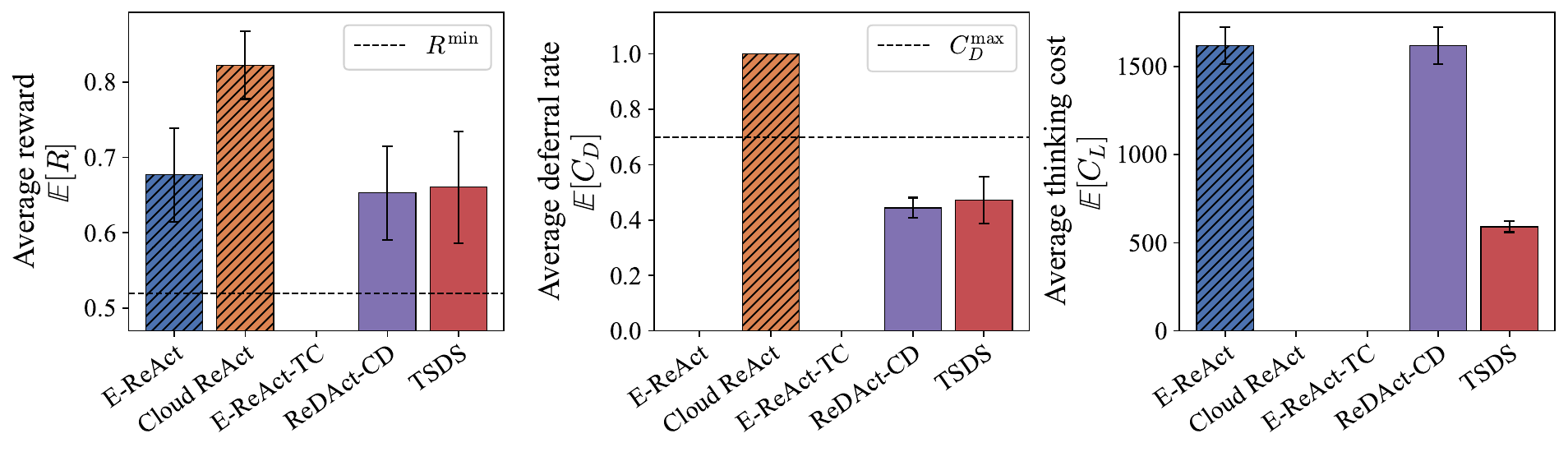}
\caption{Test results for the household robot task, shown for five policies and averaged over $50$ random $75/25$ calibration--test splits. Error bars show one standard deviation across splits.
\emph{Left}: Average reward per policy. The dashed horizontal line marks the minimum reward floor $R^{\min}=0.52$.
\emph{Center}: Average deferral rate to the cloud model. The dashed line marks the maximum allowed deferral fraction $C_D^{\max}=0.70$.
\emph{Right}: Average thinking cost (tokens).
Hatched bars denote uncalibrated reference policies.}
\label{fig:exp5_test_bars}
\end{figure}

Fig.~\ref{fig:exp5_test_bars} reports average reward, deferral rate, and thinking cost for all policies, averaged over $50$ calibration--test splits.
The overall pattern follows Figs.~\ref{fig:exp4_test_bars} and~\ref{fig:exp3_test_bars}, with ReDAct-CD and TSDS being the only two policies that satisfy both constraints, while TSDS does so at $64\%$ lower thinking cost than ReDAct-CD and E-ReAct.
One notable difference is that ReDAct-CD achieves lower reward than TSDS here, as in HotpotQA (Fig.~\ref{fig:exp3_test_bars}) and unlike in MBPP (Fig.~\ref{fig:exp4_test_bars}). This can be further evidence that full thought traces allow the edge model to grow overconfident in multi-step plans, suppressing the perplexity signal at the steps where cloud escalation is most needed, while TC-induced truncation preserves uncertainty and directs deferral to the genuinely hard steps.

\section{Conclusion}
\label{sec:conclusion}

We introduced Think Short, Defer Smart (TSDS), which composes thought calibration and uncertainty-aware deferral into a jointly calibrated pipeline for LLM agents. A convergence probe halts on-device reasoning once the intended action has stabilized, and a perplexity-based deferral rule escalates uncertain actions to a cloud model. To ensure reliability, a multi-objective Learn-Then-Test procedure certifies that reward and cloud-call rate remain within user-specified limits, selecting the operating point that minimizes thinking compute.

Across all four benchmarks, TSDS simultaneously satisfies the certified reward floor and the cloud-call budget, while consuming substantially fewer thought tokens than deferral-only baselines. This outcome highlights two complementary insights. First, thought calibration and deferral to a cloud model are synergistic, as truncated reasoning leads to detecting uncertainty signals at their most meaningful state. Second, the necessity of deferral varies sharply with task complexity. On simpler single-step tasks, truncation alone nearly suffices and deferral provides a targeted safety net, while on multi-step tasks such as the household robot task, thought calibration without deferral collapses performance severely, and cloud escalation is essential for recovery.

\paragraph{Future directions:}
Immediate extensions include richer uncertainty signals, e.g., semantic entropy,
to narrow the deferral gap toward the oracle; multi-tier cascades of agents with
multi-constraint LTT certification; and extending thought calibration to models without an explicit reasoning delimiter by
inferring action convergence directly from hidden-state trajectories
\citep{esakkiraja2026therefore}.

\section*{Acknowledgments}
This work was supported by the European Research Council (ERC) under the European Union's Horizon Europe Programme (grant agreement No.\ 101198347). The work of O. Simeone was also supported by an EPSRC Open Fellowship (EP/W024101/1) and by the EPSRC project (EP/X011852/1).

\bibliographystyle{iclr2027_conference}
\bibliography{refs}

@article{cobbe2021gsm8k,
  title   = {Training Verifiers to Solve Math Word Problems},
  author  = {Cobbe, Karl and Kosaraju, Vineet and Bavarian, Mohammad
             and Chen, Mark and Jun, Heewoo and Kaiser, Lukasz
             and Plappert, Matthias and Tworek, Jerry and Hilton, Jacob
             and Nakano, Reiichiro and Hesse, Christopher and Schulman, John},
  journal = {arXiv preprint arXiv:2110.14168},
  year    = {2021}
}

@article{kramar2026probes,
  title={Building production-ready probes for gemini},
  author={Kram{\'a}r, J{\'a}nos and Engels, Joshua and Wang, Zheng and Chughtai, Bilal and Shah, Rohin and Nanda, Neel and Conmy, Arthur},
  journal={arXiv preprint arXiv:2601.11516},
  year={2026}
}

@article{kantamneni2025attention,
  title={Are sparse autoencoders useful? a case study in sparse probing},
  author={Kantamneni, Subhash and Engels, Joshua and Rajamanoharan, Senthooran and Tegmark, Max and Nanda, Neel},
  journal={arXiv preprint arXiv:2502.16681},
  year={2025}
}

@article{austin2021program,
  title   = {Program Synthesis with Large Language Models},
  author  = {Austin, Jacob and Odena, Augustus and Nye, Maxwell
             and Bosma, Maarten and Michalewski, Henryk and Dohan, David
             and Jiang, Ellen and Cai, Carrie J. and Terry, Michael
             and Le, Quoc V. and Sutton, Charles},
  journal = {arXiv preprint arXiv:2108.07732},
  year    = {2021}
}

@inproceedings{yang2018hotpotqa,
  title={{HotpotQA}: A dataset for diverse, explainable multi-hop question answering},
  author={Yang, Zhilin and Qi, Peng and Zhang, Saizheng and Bengio, Yoshua and Cohen, William and Salakhutdinov, Ruslan and Manning, Christopher D},
  booktitle={Proceedings of the 2018 conference on empirical methods in natural language processing},
  pages={2369--2380},
  year={2018}
}

@inproceedings{yao2022react,
  title     = {{ReAct}: Synergizing Reasoning and Acting in Language Models},
  author    = {Yao, Shunyu and Zhao, Jeffrey and Yu, Dian and Du, Nan
               and Shafran, Izhak and Narasimhan, Karthik and Cao, Yuan},
  booktitle = {International Conference on Learning Representations},
  year      = {2023}
}

@inproceedings{shinn2023reflexion,
  title     = {Reflexion: Language Agents with Verbal Reinforcement Learning},
  author    = {Shinn, Noah and Cassano, Federico and Gopinath, Ashwin
               and Narasimhan, Karthik and Yao, Shunyu},
  booktitle = {Advances in Neural Information Processing Systems},
  volume    = {36},
  year      = {2023}
}

@article{wang2023voyager,
  title   = {Voyager: An Open-Ended Embodied Agent with Large Language Models},
  author  = {Wang, Guanzhi and Xie, Yuqi and Jiang, Yunfan and Mandlekar, Ajay
             and Xiao, Chaowei and Zhu, Yuke and Fan, Linxi and Anandkumar, Anima},
  journal = {arXiv preprint arXiv:2305.16291},
  year    = {2023}
}

@inproceedings{sun2023adaplan,
  title     = {{AdaPlanner}: Adaptive Planning from Feedback with Language Models},
  author    = {Sun, Haotian and Zhuang, Yuchen and Kong, Lingkai
               and Dai, Bo and Zhang, Chao},
  booktitle = {Advances in Neural Information Processing Systems},
  volume    = {36},
  year      = {2023}
}

@inproceedings{shridhar2021alfworld,
  title     = {{ALFWorld}: Aligning Text and Embodied Environments for Interactive Learning},
  author    = {Shridhar, Mohit and Yuan, Xingdi and C{\^o}t{\'e}, Marc-Alexandre
               and Bisk, Yonatan and Trischler, Adam and Hausknecht, Matthew},
  booktitle = {International Conference on Learning Representations},
  year      = {2021}
}

@inproceedings{chevalier2023minigrid,
  title     = {Minigrid \& Miniworld: Modular \& Customizable Reinforcement Learning
               Environments for Goal-Oriented Tasks},
  author    = {Chevalier-Boisvert, Maxime and Dai, Bolun and Towers, Mark
               and de Lazcano, Rodrigo and Willems, Lucas and Lahlou, Salem
               and Pal, Suman and Castro, Pablo Samuel and Terry, Jordan},
  booktitle = {Advances in Neural Information Processing Systems},
  volume    = {36},
  year      = {2023}
}

@article{zhou2024language,
  title   = {Language Agent Tree Search Unifies Reasoning, Acting, and Planning
             in Language Models},
  author  = {Zhou, Andy and Yan, Kai and Shlapentokh-Rothman, Michal
             and Wang, Haohan and Wang, Yu-Xiong},
  journal = {arXiv preprint arXiv:2310.04406},
  year    = {2024}
}

@article{malinin2021uncertainty,
  title   = {Uncertainty Estimation in Autoregressive Structured Prediction},
  author  = {Malinin, Andrey and Gales, Mark},
  journal = {arXiv preprint arXiv:2002.07650},
  year    = {2021}
}

@article{fomicheva2020unsupervised,
  title   = {Unsupervised Quality Estimation for Neural Machine Translation},
  author  = {Fomicheva, Marina and Sun, Shuo and Yankovskaya, Lisa
             and Blain, Fr{\'e}d{\'e}ric and Guzm{\'a}n, Francisco
             and Fishel, Mark and Aletras, Nikolaos
             and Chaudhary, Vishrav and Specia, Lucia},
  journal = {Transactions of the Association for Computational Linguistics},
  volume  = {8},
  pages   = {539--555},
  year    = {2020}
}

@article{chen2024frugalgpt,
  title   = {{FrugalGPT}: How to Use Large Language Models While Reducing Cost
             and Improving Performance},
  author  = {Chen, Lingjiao and Zaharia, Matei and Zou, James},
  journal = {arXiv preprint arXiv:2305.05176},
  year    = {2023}
}

@inproceedings{yue2023llm,
  title={Large language model cascades with mixture of thought representations for cost-efficient reasoning},
  author={Yue, Murong and Zhao, Jie and Zhang, Min and Du, Liang and Yao, Ziyu},
  booktitle={International Conference on Learning Representations},
  volume={2024},
  pages={21691--21728},
  year={2024}
}

@inproceedings{ong2025routellm,
  title={Routellm: Learning to route llms from preference data},
  author={Ong, Isaac and Almahairi, Amjad and Wu, Vincent and Chiang, Wei-Lin and Wu, Tianhao and Gonzalez, Joseph E and Kadous, Mohammed and Stoica, Ion},
  booktitle={International Conference on Learning Representations},
  volume={2025},
  pages={34433--34448},
  year={2025}
}

@article{piatrashyn2026redact,
  title={ReDAct: Uncertainty-Aware Deferral for LLM Agents},
  author={Piatrashyn, Dzianis and Kotelevskii, Nikita and Grishchenkov, Kirill and Glazkov, Nikita and Nasonov, Ivan and Makarov, Ilya and Baldwin, Timothy and Nakov, Preslav and Vashurin, Roman and Panov, Maxim},
  journal={arXiv preprint arXiv:2604.07036},
  year={2026}
}

@inproceedings{wu2025thought,
    title = "Thought calibration: Efficient and confident test-time scaling",
    author = "Wu, Menghua  and
      Zhou, Cai  and
      Bates, Stephen  and
      Jaakkola, Tommi",
    editor = "Christodoulopoulos, Christos  and
      Chakraborty, Tanmoy  and
      Rose, Carolyn  and
      Peng, Violet",
    booktitle = "Proceedings of the 2025 Conference on Empirical Methods in Natural Language Processing",
    month = nov,
    year = "2025",
    address = "Suzhou, China",
    publisher = "Association for Computational Linguistics",
    url = "https://aclanthology.org/2025.emnlp-main.722/",
    doi = "10.18653/v1/2025.emnlp-main.722",
    pages = "14291--14305",
    ISBN = "979-8-89176-332-6"
}

@misc{esakkiraja2026therefore,
      title={Therefore I am. I Think}, 
      author={Esakkivel Esakkiraja and Sai Rajeswar and Denis Akhiyarov and Rajagopal Venkatesaramani},
      year={2026},
      eprint={2604.01202},
      archivePrefix={arXiv},
      primaryClass={cs.AI},
      url={https://arxiv.org/abs/2604.01202}, 
}

@inproceedings{yang2026dynamic,
  title={Dynamic early exit in reasoning models},
  author={Yang, Chenxu and Si, Qingyi and Duan, Yongjie and Zhu, Zheliang and Zhu, Chenyu and Li, Qiaowei and Chen, Minghui and Lin, Zheng and Wang, Weiping},
  booktitle={International Conference on Learning Representations},
  volume={2026},
  pages={88170--88210},
  year={2026}
}

@article{angelopoulos2021learn,
  title   = {Learn then Test: Calibrating Predictive Algorithms to Achieve Risk Control},
  author  = {Angelopoulos, Anastasios N. and Bates, Stephen and
             Cand{\`e}s, Emmanuel J. and Jordan, Michael I. and Lei, Lihua},
  journal = {arXiv preprint arXiv:2110.01052},
  year    = {2021}
}

@misc{farzaneh2026statistically,
  title  = {Statistically Valid Hyperparameter Selection: From Tuning to Guarantees},
  author = {Farzaneh, Amirmohammad and Simeone, Osvaldo},
  year   = {2026},
  note   = {arXiv preprint arXiv:2606.25601}
}

@article{dunn1961multiple,
  title   = {Multiple Comparisons Among Means},
  author  = {Dunn, Olive Jean},
  journal = {Journal of the American Statistical Association},
  volume  = {56},
  number  = {293},
  pages   = {52--64},
  year    = {1961}
}

@article{guo2025deepseek,
  title   = {{DeepSeek-R1}: Incentivizing Reasoning Capability in {LLMs}
             via Reinforcement Learning},
  author  = {Guo, Daya and Yang, Dejian and Zhang, Haowei and Song, Junxiao
             and Zhang, Ruoyu and Xu, Runxin and Zhu, Qihao and Ma, Shirong
             and Wang, Peiyi and Bi, Xiao and others},
  journal = {arXiv preprint arXiv:2501.12948},
  year    = {2025}
}

@article{ahn2022saycan,
  title   = {Do As {I} Can, Not As {I} Say: Grounding Language in Robotic Affordances},
  author  = {Ahn, Michael and Brohan, Anthony and Brown, Noah and others},
  journal = {arXiv preprint arXiv:2204.01691},
  year    = {2022}
}

@article{xu2024ondevice,
  title={On-device language models: A comprehensive review},
  author={Xu, Jiajun and Li, Zhiyuan and Chen, Wei and Wang, Qun and Gao, Xin and Cai, Qi and Ling, Ziyuan},
  journal={arXiv preprint arXiv:2409.00088},
  year={2024}
}

@article{hou2026reliable,
  title={Reliable LLM-Based Edge-Cloud-Expert Cascades for Telecom Knowledge Systems},
  author={Hou, Qiushuo and Park, Sangwoo and Zecchin, Matteo and Cai, Yunlong and Yu, Guanding and Simeone, Osvaldo and Melodia, Tommaso},
  journal={IEEE Transactions on Communications},
  year={2026},
  publisher={IEEE}
}

@article{chen2024knowing,
  title={Knowing when to stop: Delay-adaptive spiking neural network classifiers with reliability guarantees},
  author={Chen, Jiechen and Park, Sangwoo and Simeone, Osvaldo},
  journal={IEEE Journal of Selected Topics in Signal Processing},
  volume={19},
  number={1},
  pages={88--102},
  year={2024},
  publisher={IEEE}
}

@article{hoeffding1963probability,
  title={Probability inequalities for sums of bounded random variables},
  author={Hoeffding, Wassily},
  journal={Journal of the American statistical association},
  volume={58},
  number={301},
  pages={13--30},
  year={1963},
  publisher={Taylor \& Francis}
}

@inproceedings{muennighoff2025s1,
  title={s1: Simple test-time scaling},
  author={Muennighoff, Niklas and Yang, Zitong and Shi, Weijia and Li, Xiang Lisa and Fei-Fei, Li and Hajishirzi, Hannaneh and Zettlemoyer, Luke and Liang, Percy and Cand{\`e}s, Emmanuel and Hashimoto, Tatsunori B},
  booktitle={Proceedings of the 2025 Conference on Empirical Methods in Natural Language Processing},
  pages={20286--20332},
  year={2025}
}

\clearpage
\appendix

\section{Code Availability}
\label{app:code}

The full implementation of TSDS, including rollout collection, convergence probe training, LTT calibration, and all experiment scripts, is available at:
\begin{center}
\url{https://github.com/amirfar76/think-short-defer-smart}
\end{center}
The Household Robot Task benchmark environment introduced in Sec.~\ref{sec:exp_household} is released as a standalone package at:
\begin{center}
\url{https://github.com/amirfar76/household-robot-bench}
\end{center}

\section{Uncertainty Measures}
\label{app:uncertainty}

The uncertainty score $u_t$ in the deferral rule (Sec.~\ref{sec:background}) can be any
information-theoretic quantity derived from the token-level probabilities of the generated
action sequence $y = (\tau_1, \ldots, \tau_L)$ given context $x$, with joint probability
\begin{equation}
p(y \mid x, \theta) = \prod_{i=1}^{L} p(\tau_i \mid x, \tau_{<i}, \theta).
\end{equation}
All three measures below are available as free by-products of the autoregressive decoding
pass, requiring no additional forward pass \citep{malinin2021uncertainty}.

\paragraph{Sequence Probability (SP):}
\begin{equation}
U_{\mathrm{SP}}(y \mid x) = -\sum_{i=1}^{L} \log p(\tau_i \mid x, \tau_{<i}, \theta).
\label{eq:sp}
\end{equation}
Following the terminology of \citet{piatrashyn2026redact}, we refer to
this score as Sequence Probability (SP). More precisely, it is the
negative log-probability of the sequence. Because it sums the
token-level negative log-likelihoods, the score is length-sensitive:
longer sequences tend to receive higher values even when their average
per-token confidence is unchanged.

\paragraph{Perplexity (PPL):}
\begin{equation}
U_{\mathrm{PPL}}(y \mid x) = -\frac{1}{L}\sum_{i=1}^{L} \log p(\tau_i \mid x, \tau_{<i}, \theta).
\label{eq:ppl}
\end{equation}
Following the terminology of \citet{piatrashyn2026redact}, we refer to
this score as Perplexity (PPL). More precisely, it is the
log-perplexity, with conventional perplexity given by
$\exp\!\left(U_{\mathrm{PPL}}(y\mid x)\right)$. Because it averages the
token-level negative log-likelihoods, it removes the direct dependence
on sequence length. Higher values indicate lower average confidence in
the generated tokens. This is the primary uncertainty score used in
our experiments.

\paragraph{Mean Token Entropy (MTE):}
\begin{equation}
U_{\mathrm{MTE}}(y \mid x) = \frac{1}{L}\sum_{i=1}^{L} H(\tau_i \mid x, \tau_{<i}, \theta),
\label{eq:mte}
\end{equation}
where $H(\tau_i \mid \cdot)$ is the Shannon entropy of the token distribution at position $i$.
We refer to this score as Mean
Token Entropy (MTE), consistent with the standard definition. Whereas
SP and PPL depend only on the probabilities assigned to the generated
tokens, MTE measures the average uncertainty of the full next-token
distribution. It can therefore capture uncertainty that SP and PPL may
miss when the generated token has high probability but substantial
probability mass remains distributed across alternative tokens \citep{malinin2021uncertainty}.

\section{Learn-Then-Test Calibration Procedure}
\label{app:ltt}

For each candidate $\boldsymbol{\lambda}_k \in \Lambda$, we define two
null hypotheses corresponding to constraint violations: the \emph{reward null}
$\mathcal{H}_k^R : R(\boldsymbol{\lambda}_k) \leq R^{\min}$, i.e., that the reward constraint is
violated, and the \emph{cost null} $\mathcal{H}_k^C : C_D(\boldsymbol{\lambda}_k) \geq
C_D^{\max}$, i.e., that the deferral budget is exceeded.
With $\varphi_\theta$ and $\Lambda$ fixed before $\mathcal{D}_{\mathrm{cal}}$ is observed, we compute
one-sided p-values $p_k^R$ and $p_k^C$ for the null hypotheses $\mathcal{H}_k^R$ and $\mathcal{H}_k^C$, respectively.
For each candidate $\boldsymbol{\lambda}_k$, the p-values are computed from the calibration data $\mathcal{D}_{\mathrm{cal}}(\boldsymbol{\lambda}_k)$ collected under that specific pair (Sec.~\ref{sec:multi_obj_calibration}), so that each p-value is evaluated on data generated by exactly the policy under test. When rewards are binary, the exact one-sided binomial p-value for the
null hypothesis $H_k^R:R(\lambda_k)\leq R^{\min}$ is
\begin{equation}
p_k^R
=
\Pr\!\left[
\mathrm{Binomial}(N,R^{\min})
\geq
\left\lfloor N\bar r_k\right\rfloor
\right],
\label{eq:pR_binom}
\end{equation}
where $\bar r_k$ is the sample mean reward. This is the probability that
$N$ i.i.d.\ Bernoulli trials with success probability $R^{\min}$
produce at least as many successes as observed.

For rewards that are not binary but are bounded in $[0,1]$, a valid
distribution-free p-value for $H_k^R$ is
\begin{equation}
p_k^R
=
\exp\!\left(
-2N\bigl[\bar r_k-R^{\min}\bigr]_+^2
\right),
\label{eq:pR_hoeffding}
\end{equation}
where $[x]_+=\max\{x,0\}$ \citep{hoeffding1963probability}.
Since the episode-level deferral rates are also bounded in $[0,1]$, a
valid distribution-free p-value for the null hypothesis
$H_k^C:C_D(\lambda_k)\geq C_D^{\max}$ is
\begin{equation}
p_k^C
=
\exp\!\left(
-2N\bigl[C_D^{\max}-\bar c_k\bigr]_+^2
\right),
\label{eq:pC_hoeffding}
\end{equation}
where $\bar c_k$ is the sample mean episode-level deferral rate.
Since the statistics $p_k^R$ and $p_k^C$ are valid p-values for their respective nulls
\citep{angelopoulos2021learn}, the maximum $p_k = \max(p_k^R, p_k^C)$ is a valid p-value
for the union null $\mathcal{H}_k^R \cup \mathcal{H}_k^C$ \citep{angelopoulos2021learn}.  The certified set is then
\begin{equation}
\hat{\Lambda} = \mathcal{A}\!\bigl(\{p_k\},\, \delta\bigr), \label{eq:certified_set}
\end{equation}
where $\mathcal{A}$ is any FWER-controlling procedure \citep{dunn1961multiple}.

We select the final hyperparameter $\hat{\boldsymbol{\lambda}}$ from the certified set $\hat{\Lambda}$ via lexicographic minimization over $\hat{\Lambda}$. We
first minimize the thinking cost $C_L(\boldsymbol{\lambda})$ \eqref{eq:deferral_cost_bg}, and in multi-step
settings, break ties by minimizing the episode-length cost $C_S(\boldsymbol{\lambda})$. Any
remaining ties are broken by minimizing $C_D(\boldsymbol{\lambda})$.

We adopt \textit{Bonferroni correction} \citep{dunn1961multiple} as the default choice for $\mathcal{A}$ as it requires no assumptions on the structure of $\Lambda$, and provides unconditional FWER control in any setting.

\section{Thought Calibration: Stopping Rule Illustration}
\label{app:algorithm}

\begin{figure}[h]
\centering
\includegraphics[width=\textwidth]{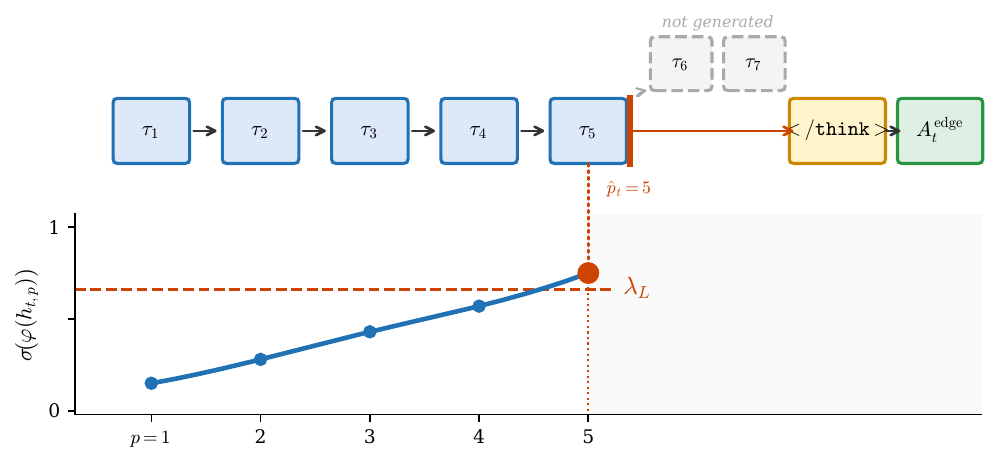}
\caption{Illustration of the thought calibration stopping rule.  The probe signal
$\sigma(\varphi_\theta(h_{t,i}))$ rises as the intended action stabilizes.  Generation halts at
the first position crossing the certified threshold $\lambda_L$ (here $i=5$),
leaving $\tau_{t,6}, \tau_{t,7}$ ungenerated.  The delimiter \texttt{</think>} is injected
and the model generates $A_t^\mathrm{edge}$ from the partial trace.}
\label{fig:probe}
\end{figure}

Fig.~\ref{fig:probe} illustrates the stopping rule from Section~\ref{sec:probe} on a single interaction step.
The edge model generates thought tokens $\tau_{t,1}, \tau_{t,2}, \ldots$ sequentially; at each strided position $i \in \mathcal{P}_t$, the convergence probe $\varphi_\theta$ evaluates the hidden state $h_{t,i}^{(\ell)}$ and returns a scalar score.
The score rises as the model's intermediate action stabilises: early in the thought, the model is still exploring reasoning paths and its committed action fluctuates (score near zero); as reasoning consolidates, the score climbs toward one.

Stopping fires the first time the probe score crosses the certified threshold $\lambda_L$.
In the example shown, this occurs at position $i = 5$: tokens $\tau_{t,6}$ and $\tau_{t,7}$ are never generated, saving two of the seven allocated positions.
The delimiter \texttt{</think>} is injected immediately after $\tau_{t,1:5}$, and the edge model generates $A_t^\mathrm{edge}$ from the truncated trace.
The action is then passed to the deferral mechanism (Section~\ref{sec:deferral_mechanism}): its PPL score is compared to $\lambda_D$, and if the action is insufficiently confident the cloud model is invoked instead.

\label{app:tc_fig}

\section{Probe Architecture Details}
\label{app:probes}

We compare seven candidate instantiations of the convergence probe $\varphi_\theta$, all trained
by minimizing the binary cross-entropy loss with soft targets using AdamW (learning rate
$10^{-3}$, weight decay $10^{-3}$, full-batch, early stopping on validation loss with patience
$20$ out of a maximum $200$ epochs).  Window probes use the $W = 16$ most recent strided
probe positions left-padded with zeros and a binary mask.

\begin{itemize}
\item \textbf{Linear:} logistic regression $\varphi(h) = w^\top h + b$ on the single hidden
vector $h_{t,p}^{(\ell)}$.
\item \textbf{MLP:} two-layer MLP with hidden width $100$ and ReLU activations on the single
vector $h_{t,p}^{(\ell)}$.
\item \textbf{Mean-linear:} per-position linear score $u_j = w^\top x_j + b$ averaged over
unmasked window slots \citep[\S3.1.1]{kramar2026probes}.
\item \textbf{EMA:} trained identically to Mean-linear; at inference, per-position scores are
passed through an EMA ($e_j = \alpha u_j + (1-\alpha)e_{j-1}$, $\alpha = 0.5$) and the probe
output is $\max_j e_j$ \citep[\S3.1.2]{kramar2026probes}.
\item \textbf{Attention probe:} per-position MLP transform followed by softmax-weighted
aggregation across $K = 4$ heads, aggregated by a linear projection
\citep{kantamneni2025attention}.
\item \textbf{MultiMax probe:} same per-position MLP transform but with hard $\max$ over the
window for $K = 4$ heads, aggregated by a linear projection
\citep[\S3.2.1]{kramar2026probes}.
\item \textbf{Rolling-mean attention probe:} per-position MLP transform, attention-weighted
mean over each sub-window of width $w' = 4$ strided positions, maximum over $K = 10$
sub-window scores.  The architecture identified as strongest across a broad range of probing
tasks in \citep[\S3.2.2]{kramar2026probes}.
\end{itemize}

\section{Proof of Proposition~\ref{prop:joint}}
\label{app:proof}

The p-values $p_k^R$ \eqref{eq:pR_binom} and $p_k^C$ are valid one-sided p-values for
$\mathcal{H}_k^R$ and $\mathcal{H}_k^C$ respectively \citep{angelopoulos2021learn}; hence
$p_k = \max(p_k^R, p_k^C)$ is a valid p-value for the joint null
$\mathcal{H}_k^R \cup \mathcal{H}_k^C$.  Because $\mathcal{A}$ controls the FWER at level
$\delta$, the probability that any true joint null is rejected is at most $\delta$, which is
equivalent to \eqref{eq:joint_guarantee}.

\section{Additional Experimental Details}
\label{app:exp_details}

\subsection{Convergence Probe Ablation}
\label{app:probe_details}

We compare seven probe architectures on $N_{\mathrm{ep}} = 100$ GSM8K train-split problems
with \texttt{DeepSeek-R1-Distill-Qwen-7B} ($28$ layers, $d = 3584$, greedy, bfloat16),
yielding $1076$ probe samples at stride $16$ (cap $L_{\max} = 384$).
The probe target is the \emph{stable-run label},
\begin{equation}
y_{t,p} \;=\; \prod_{p' \in \mathcal{P}_t,\, p' \geq p}\mathbf{1}\bigl[A_{p',t} = A_t\bigr],
\label{eq:stable_label}
\end{equation}
which is $1$ iff the early-stopped action agrees with the full-thought action at position $p$
\emph{and} at every subsequent probe position in step $t$, directly aligned with the
irrevocability of the stopping rule (Sec.~\ref{sec:probe}).
To isolate hidden-state signal from the global trend in positive rate across the thought
trace, we stratify by relative position $p/L_t$ and report per-quartile AUROC
\begin{equation}
\mathrm{AUROC}_{Q_q} = \mathrm{AUROC}\!\bigl(\{(\hat{s}_i, y_i) : p_i/L_{t_i} \in
  [(q{-}1)/4,\, q/4)\}\bigr), \quad q \in \{1, 2, 3, 4\}.
\label{eq:auroc_q}
\end{equation}

Windowed probes outperform single-state probes: the EMA probe leads with pooled AUROC
$0.815$ at $\ell^\star=5$ (vs.\ ${\leq}\,0.673$ for single-state), generalizing consistently
across all four quartiles.  Best probing layers are shallow ($\ell \approx 5$ for windowed
probes; $\ell = 16$ for single-state), suggesting action-relevant information concentrates
early in the transformer stack and benefits from temporal smoothing across the window.
The non-monotone quartile pattern (Q2 highest at $0.857$, Q3 dip at $0.748$, Q4 recovery
to $0.812$ for EMA) reflects class-balance shifts: Q1 has a low positive rate (stability
requires all subsequent probe positions to agree, rare for $p \ll L_t$); Q2 shows peak
discrimination because the model encodes the action before committing irrevocably; Q4
recovers as the model converges toward its natural stopping string.

\begin{table}[h]
\centering
\caption{Probe ablation on GSM8K with \texttt{DeepSeek-R1-Distill-Qwen-7B}
($N_{\mathrm{ep}} = 100$ problems, $1076$ probe samples, $3$ seeds) under the stable-run
label \eqref{eq:stable_label}.  For each probe architecture we report the best layer
$\ell^\star$ and its seed-averaged pooled AUROC, plus per-quartile AUROC
$\mathrm{AUROC}_{Q_1}, \ldots, \mathrm{AUROC}_{Q_4}$ (defined in \eqref{eq:auroc_q}).
Bold marks the best entry in each column.}
\label{tab:exp1_arch}
\small
\begin{tabular}{l c c c c c c}
\toprule
Probe & $\ell^\star$ & pooled AUROC & $Q_1$ & $Q_2$ & $Q_3$ & $Q_4$ \\
\midrule
Linear              & $16$ & $0.637$          & $0.494$          & $0.688$          & $0.606$          & $0.603$ \\
MLP                 & $16$ & $0.673$          & $0.496$          & $0.727$          & $0.606$          & $0.713$ \\
Mean-linear         & $5$  & $0.686$          & $0.608$          & $0.687$          & $0.663$          & $0.758$ \\
Attention           & $5$  & $0.733$          & $0.672$          & $0.727$          & $0.708$          & $0.823$ \\
MultiMax            & $26$ & $0.626$          & $0.556$          & $0.700$          & $0.725$          & $0.554$ \\
Rolling-mean attn.  & $5$  & $0.762$          & $0.672$          & $0.770$          & $0.733$          & $\mathbf{0.874}$ \\
\textbf{EMA}        & $5$  & $\mathbf{0.815}$ & $\mathbf{0.817}$ & $\mathbf{0.857}$ & $\mathbf{0.748}$ & $0.812$ \\
\bottomrule
\end{tabular}
\end{table}

\paragraph{Reasoning model and data collection:}
We use \texttt{DeepSeek-R1-Distill-Qwen-7B} with $28$ transformer layers and hidden size
$d = 3584$.  For each of the $100$ GSM8K train-split problems, we decode the thought trace
greedily until \texttt{</think>} or $L_{\max} = 384$ tokens.  A single forward pass with
hidden-state output provides $h_{t,p}^{(\ell)}$ at every thought position and every layer at
zero marginal cost.  For each strided position $p \in \mathcal{P}_t$, we splice
\texttt{</think>} after $\tau_{t,1:p}$ and decode to obtain $A_{p,t}$.  The convergence label
$y_{t,p}$ is then computed via \eqref{eq:stable_label}.

\paragraph{Ablation grid and metrics:}
We train each of the seven architectures at every layer $\ell \in \{1,\ldots,28\}$ from three
random initializations with the 60/20/20 problem-level split fixed, giving $7 \times 28 \times
3 = 588$ configurations.  The formal AUROC is the U-statistic
\begin{equation}
\mathrm{AUROC} = \Pr\!\bigl(\hat{s}_I > \hat{s}_J \;\big|\; y_I = 1,\; y_J = 0\bigr), \quad
I, J \sim \mathrm{Uniform}\{1,\ldots,n\}\ \text{independently},
\label{eq:auroc}
\end{equation}
equal to $0.5$ for a probe whose scores are independent of the label.

\begin{figure}[t]
\centering
\includegraphics[width=0.75\textwidth]{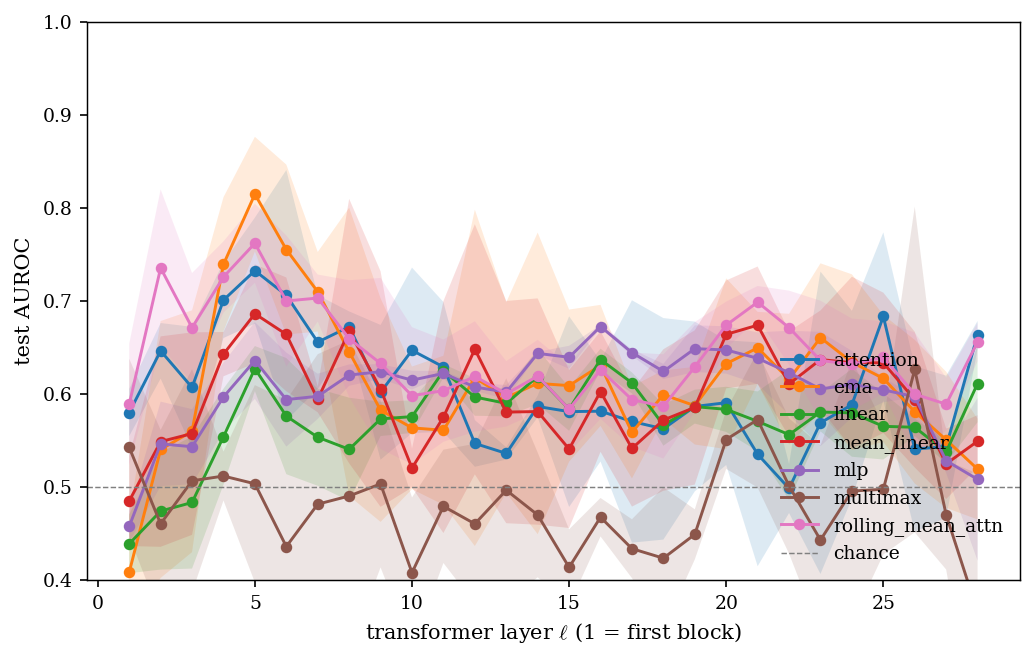}
\caption{Test AUROC of $\varphi_\theta(h_{t,p}^{(\ell)})$ vs.\ probing layer $\ell$ on GSM8K
under the stable-run label \eqref{eq:stable_label}.  Solid lines: seed-averaged test
AUROC; shaded bands: $\pm 1$ standard deviation.  Dashed line: chance ($0.5$).  The EMA and
rolling-mean attention probes dominate at $\ell \approx 5$.}
\label{fig:exp1_auroc_layer}
\end{figure}

\begin{figure}[t]
\centering
\includegraphics[width=0.75\textwidth]{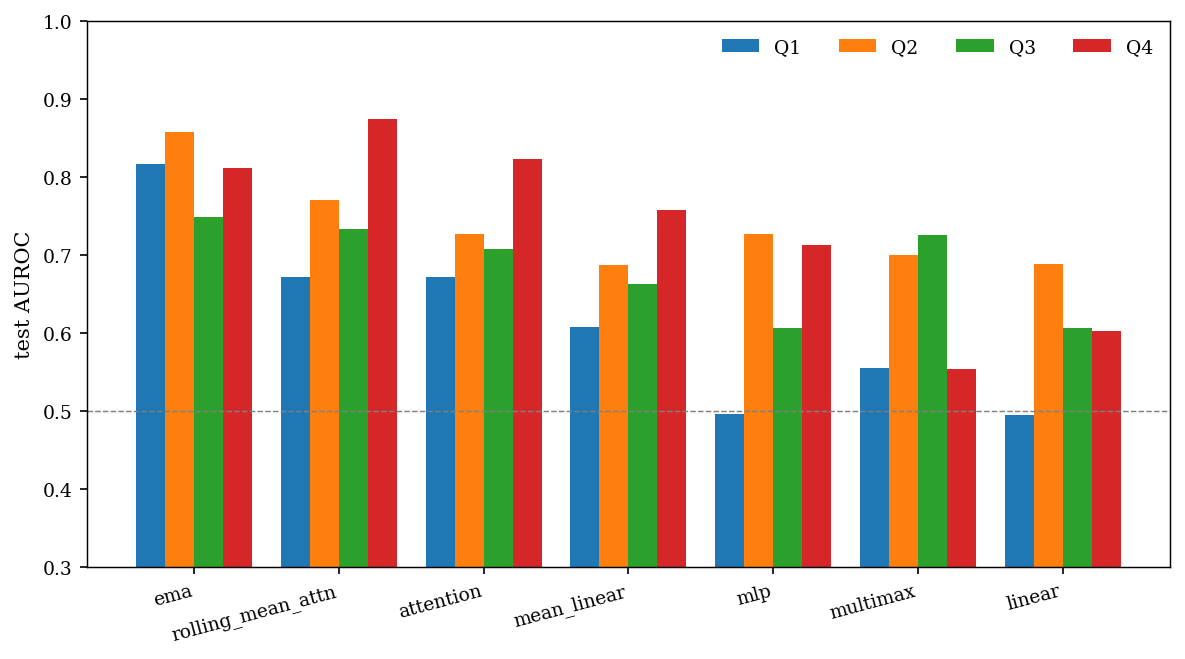}
\caption{Per-quartile AUROC at the best layer $\ell^\star$ for each probe architecture on
GSM8K under the stable-run similarity score \eqref{eq:stable_label}.  The EMA probe leads
on $Q_1$, $Q_2$, and $Q_3$, while the rolling-mean attention probe attains the highest
late-thought AUROC ($\mathrm{AUROC}_{Q_4} = 0.874$).  The dashed grey line marks chance
performance ($\mathrm{AUROC} = 0.5$).}
\label{fig:exp1_auroc_quartile}
\end{figure}

\paragraph{Position-only floor:}
A logistic regression on the relative thought position $p/L_t$ alone achieves
$\mathrm{AUROC} = 0.530$.  This \emph{position-only floor} quantifies the trivial signal
available from late-thought samples having higher positive rates.  Only $22\%$ of problems
are stable throughout their thought; the median first-stable position falls at $91\%$ of
thought length, and the mean flip count is $2.8$ per step.

\subsection{GSM8K End-to-End TSDS}
\label{app:gsm8k_details}

We evaluate TSDS on GSM8K \citep{cobbe2021gsm8k}, a benchmark of grade-school arithmetic word problems.
The task is single-step ($T=1$): the agent generates a complete reasoning trace followed by a final numerical answer in a boxed format, and reward is a binary exact-match indicator.

\paragraph{Setup:}
We use \texttt{DeepSeek-R1-Distill-Qwen-7B} as the edge model $\pi^\mathrm{edge}$
($L_{\max}=384$, mean thought length $159$ tokens) and
\texttt{DeepSeek-R1-Distill-Qwen-14B} as the cloud agent $\pi^\mathrm{cloud}$
($L_{\max}=1024$, mean thought length $741$ tokens).
The EMA probe is set at layer $\ell=5$ and trained on $100$ GSM8K train-split problems.
We evaluate on $1300$ held-out problems averaged over $50$ random $85/15$ calibration--test
splits ($N_{\mathrm{test}}=195$ per split); full-thought accuracies are $84.5\%$ for
$\pi^\mathrm{edge}$ and $90.2\%$ for $\pi^\mathrm{cloud}$.

\paragraph{Calibration:}
The reasoning-threshold grid is $\lambda_L \in \{0.1, 0.2, \ldots, 0.9\}$.
The deferral grid $\lambda_D \in \{0.053, 0.063, 0.073, 0.088, 0.127\}$ [nat] is set to
quantiles $30$--$95$ of the PPL distribution on $\mathcal{D}_{\mathrm{tr}}$, following the
general procedure described in Sec.~\ref{sec:exp_hotpotqa}.
This gives $|\Lambda|=45$ candidates.
We use exact one-sided binomial p-values with Bonferroni correction at $\delta=0.10$,
$R^{\min}=0.855$, and $C_D^{\max}=0.70$.

\paragraph{Results:}

\begin{table}[h]
\centering
\caption{GSM8K test-set results ($\mathcal{D}_{\mathrm{test}}$), means over $50$ random splits; each policy evaluated at its own LTT-selected $\hat{\boldsymbol{\lambda}}$.
$\hat{R}$: empirical reward; $\hat{C}_D$: deferral rate; $\hat{C}_L$: mean thinking tokens.
Hatched entries fail the joint LTT constraints.
Bold: lowest $\hat{C}_L$ among policies satisfying the joint constraints.}
\label{tab:gsm8k_test}
\small
\begin{tabular}{l c c c}
\toprule
Policy & $\hat{R}$ & $\hat{C}_D$ & $\hat{C}_L$ (tok) \\
\midrule
\multicolumn{4}{l}{\textit{GSM8K} ($n_{\mathrm{test}}{=}195$);
  $\hat{\boldsymbol{\lambda}}{=}(0.10,\,0.088)$, $R^{\min}{=}0.855$, $C_D^{\max}{=}0.70$} \\[2pt]
E-ReAct                       & $0.852$ & $0.000$ & $173$ \\
Cloud ReAct                   & $0.906$ & $1.000$ & $\text{---}$ \\
E-ReAct-TC ($\lambda_L{=}0.10$) & $0.852$ & $0.000$ & $\phantom{00}18$ \\
ReDAct-CD ($\lambda_L{=}\infty$)   & $0.864$ & $0.082$ & $173$ \\
\textbf{TSDS @ $\hat{\boldsymbol{\lambda}}$} & $\mathbf{0.890}$ & $0.255$ & $\phantom{00}\mathbf{18}$ \\
\bottomrule
\end{tabular}
\end{table}

\begin{figure}[h]
\centering
\includegraphics[width=\textwidth]{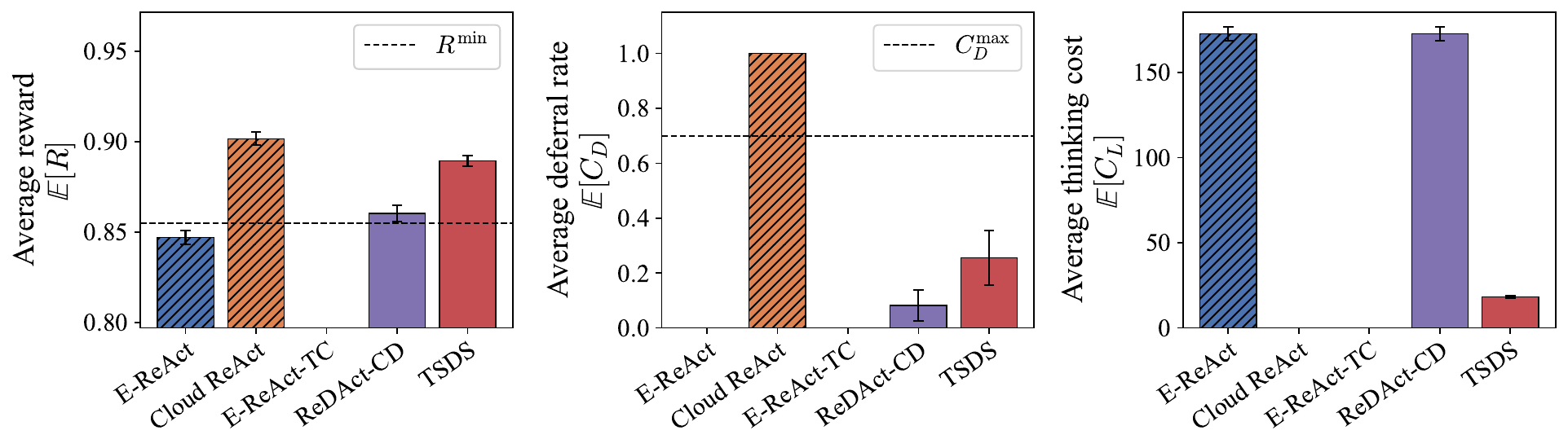}
\caption{GSM8K test-set results ($50$-split averages $\pm$ std, $n_{\mathrm{test}}{=}195$ per split, $\delta=0.10$).
Across $50$ independent $85/15$ splits of $1300$ held-out problems, each method's operating
point $\hat{\boldsymbol{\lambda}}$ is selected by LTT on the calibration portion; all metrics
are measured on the held-out test portion.
\emph{Left}: test-set reward; dashed line marks $R^{\min}=0.855$.
A bar above this line reflects empirical test performance, not the LTT guarantee.
\emph{Center}: test-set deferral rate, $C_D^{\max}=0.70$ dashed.
\emph{Right}: mean thinking tokens.
The E-ReAct-TC bar is empty because it consistently falls below the reward floor $R^{\min}$; its empirical values are listed in the table above for reference.
TSDS and ReDAct-CD satisfy the joint constraints.}
\label{fig:exp2_test_bars}
\end{figure}

TSDS reaches $\hat{R}_{\mathrm{test}}=0.890$, a $0.026$ improvement over ReDAct-CD, while using just $18$ thought tokens per problem, one-tenth of ReDAct-CD's cost.
ReDAct-CD retains the full thought trace ($173$ tokens) and defers on only $8\%$ of problems, achieving $\hat{R}_{\mathrm{test}}=0.864$.
E-ReAct-TC does not certify: without a deferral safety net, truncation-induced errors have no recovery path and the reward lower bound consistently fails to clear the floor.
The elevated deferral rate of TSDS ($26\%$) relative to ReDAct-CD ($8\%$) reflects the thought-calibration--deferral synergy: aggressively truncated thoughts produce higher PPL, directing more problems to $\pi^\mathrm{cloud}$, with that deferral concentrated on the problems where truncated reasoning is most fragile.
Cloud ReAct achieves $\hat{R}_{\mathrm{test}}=0.906$ but defers unconditionally ($C_D=1.0$), violating $C_D^{\max}=0.70$.

\paragraph{Rollout quality:}
On $1300$ test-split problems, full-thought accuracy is $84.5\%$ for $\pi^\mathrm{edge}$
and $90.2\%$ for $\pi^\mathrm{cloud}$.  The two models agree on $1026$ problems, are both
wrong on $54$, only $\pi^\mathrm{edge}$ is right on $73$, and only $\pi^\mathrm{cloud}$
is right on $147$.  An oracle deferral catching exactly the $147$ ``only-large-right''
problems would lift accuracy from $84.5\%$ to $95.8\%$.

\paragraph{Baseline details:}
\begin{itemize}
\item \emph{E-ReAct} ($\lambda_L = \lambda_D = +\infty$): full thought, no deferral.
\item \emph{Cloud ReAct} ($\lambda_D = -\infty$): always defer; no Thought Calibration.
\item \emph{E-ReAct-TC} ($\lambda_D = +\infty$, $\lambda_L = 0.10$): truncate but never
defer; excluded from $\hat{\Lambda}$ because $\hat{R}_{\mathrm{cal}} = 0.839 < R^{\min}$.
\item \emph{ReDAct-CD} ($\lambda_L = \infty$, $\lambda_D = 0.088$): full thought; $\lambda_D$
selected via single-objective LTT; certifies on reward.
\end{itemize}

\subsection{HotpotQA: Bar Chart and Pareto Frontier}
\label{app:hotpotqa_details}

\begin{figure}[t]
\centering
\includegraphics[width=\textwidth]{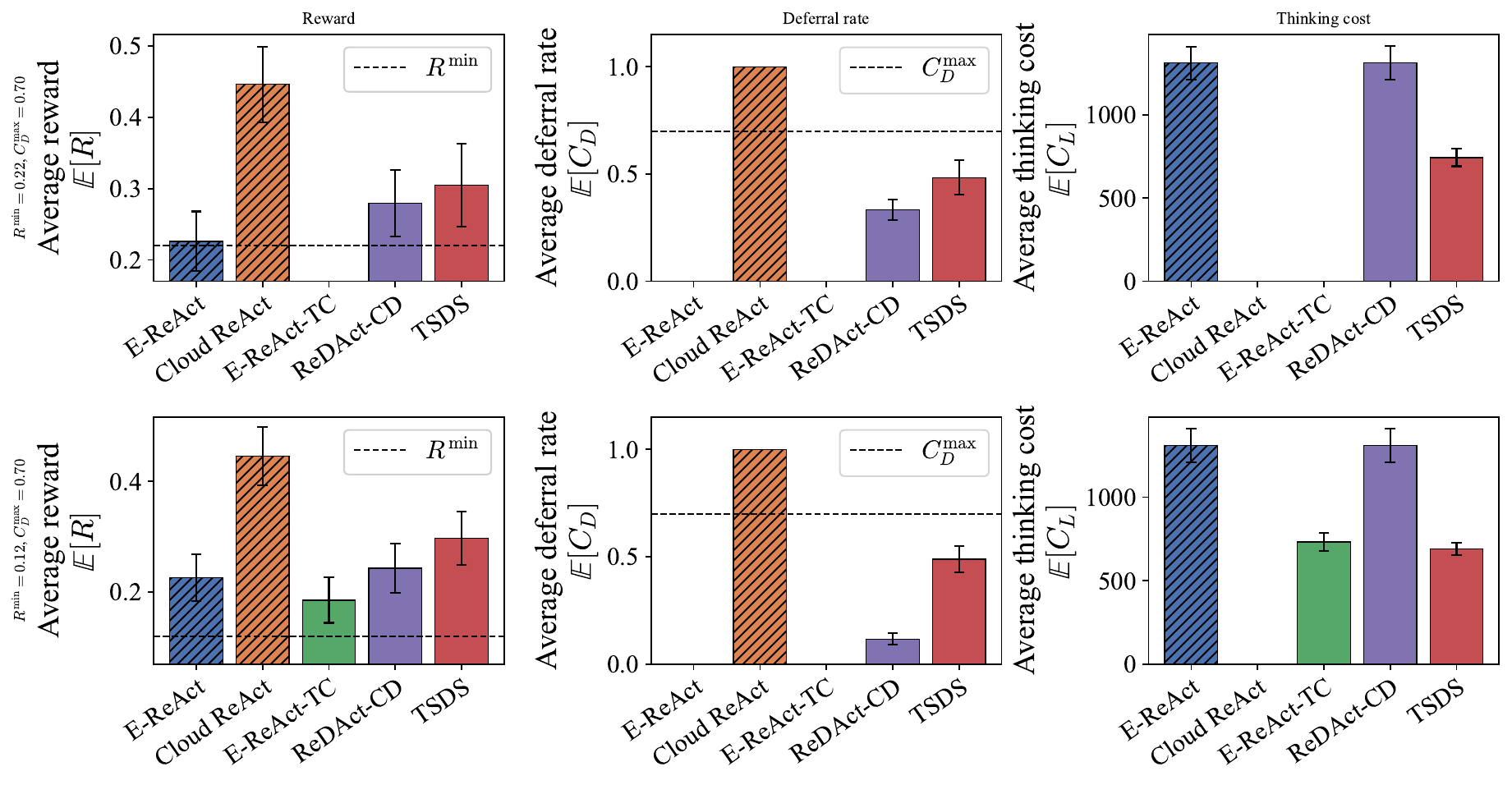}
\caption{HotpotQA results under two constraint regimes, averaged over $50$ random splits.
Each panel shows test-set means and $\pm 1$ std across $50$ independent $75/25$ splits
of $250$ episodes; the $R^{\min}$ and $C_D^{\max}$ lines mark the bounds
derived from $\mathcal{D}_{\mathrm{cal}}$, not thresholds on test-set values.
\emph{Upper} ($R^{\min}=0.22$): E-ReAct-TC ($\hat{R}_{\mathrm{cal}}{=}0.202$) and
E-ReAct ($\hat{R}_{\mathrm{cal}}{=}0.218$) fall below the reward floor;
Cloud ReAct exceeds $C_D^{\max}$; TSDS and ReDAct-CD satisfy the joint constraints.
\emph{Lower} ($R^{\min}=0.12$): E-ReAct-TC also satisfies the reward constraint; TSDS selects a more aggressive
$\lambda_L{=}0.10$, further reducing thinking cost with the deferral safety net.
Columns: $\hat{R}$, $\hat{C}_D$, $\hat{C}_L$.  Hatched bars fail the joint constraints.}
\label{fig:exp3_combined_bars}
\end{figure}

\paragraph{Bar-chart comparison:}
Fig.~\ref{fig:exp3_combined_bars} shows how the selected policies change when the reward requirement is relaxed. Under the stricter reward floor $R^{\min}=0.22$, only TSDS and ReDAct-CD satisfy both constraints; compared with ReDAct-CD, TSDS achieves higher test reward while using approximately $43\%$ fewer thinking tokens ($742$ versus $1312$), with its deferral rate remaining below $C_D^{\max}=0.70$. When the reward floor is lowered to $R^{\min}=0.12$, E-ReAct-TC also becomes feasible, and TSDS selects the more aggressive threshold $\lambda_L=0.10$, further reducing its thinking cost to $691$ tokens while continuing to satisfy both constraints. These results illustrate how joint calibration adapts the operating point to the specified reward requirement rather than committing to a fixed reasoning policy.

\paragraph{Pareto frontier:}

\begin{figure}[t]
\centering
\includegraphics[width=0.85\textwidth]{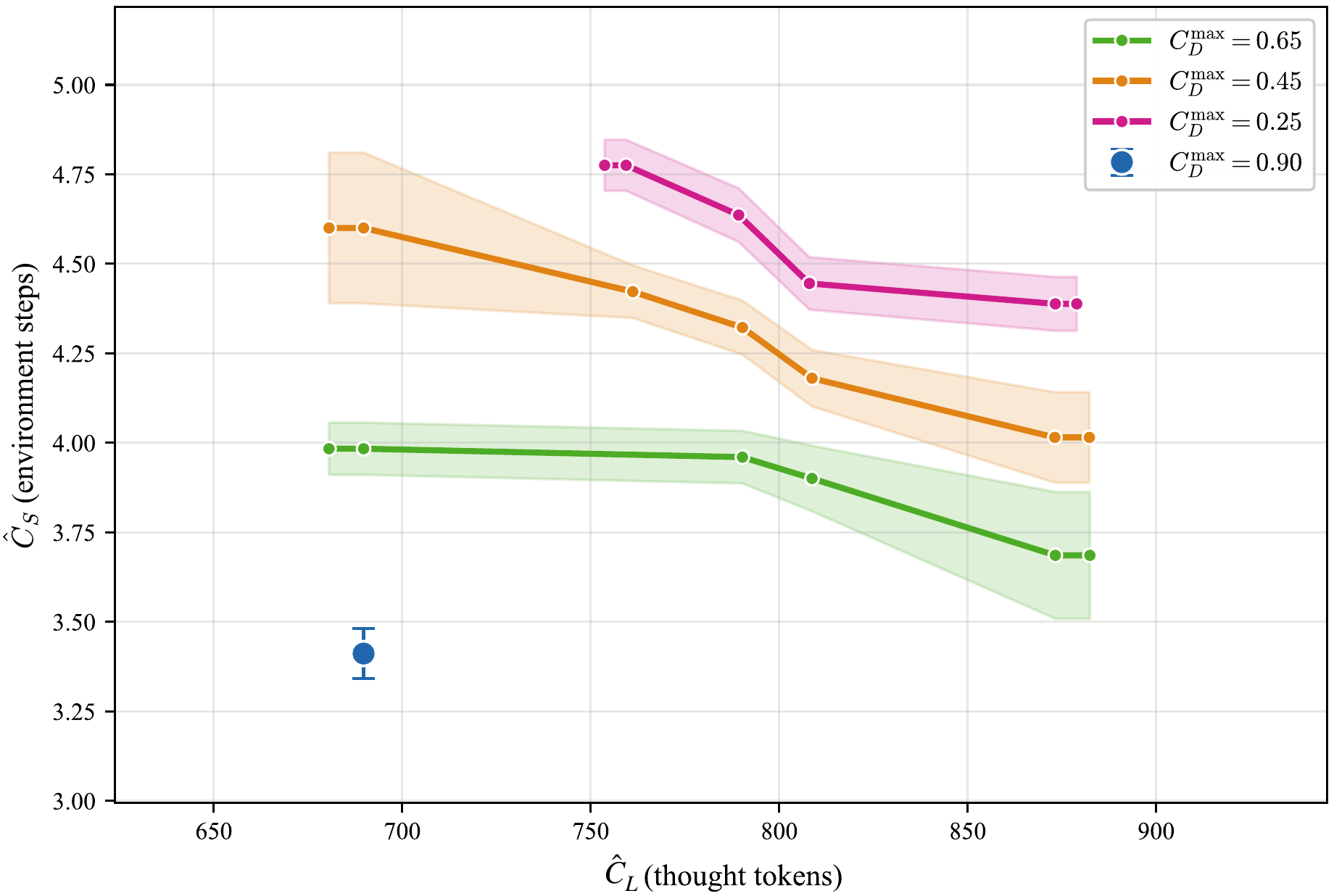}
\caption{Pareto frontiers in $(\hat{C}_L, \hat{C}_S)$ space on HotpotQA for four deferral
budgets $C_D^{\max}$, averaged over $200$ random
75/25 calibration/test splits.  For each split and each certified $\lambda_L$, we record
$(\hat{C}_L,\, \min_{\lambda_D \in \hat{\Lambda}(\lambda_L)} \hat{C}_S)$; plotted points
are means across splits; shaded bands are $\pm 1$ std of $\hat{C}_S$.}
\label{fig:exp3_pareto}
\end{figure}

Fig.~\ref{fig:exp3_pareto} shows how the Pareto trade-off between thinking cost and episode length changes with the cloud-deferral budget. Each curve corresponds to a fixed value of $C_D^{\max}$ and contains the Pareto-optimal certified operating points obtained by varying $\lambda_L$. Within a curve, moving left corresponds to more aggressive thought truncation: it reduces $\hat{C}_L$ but increases $\hat{C}_S$, meaning that fewer thought tokens come at the cost of longer interaction trajectories. Across curves, relaxing the deferral budget shifts the frontier down and to the left, because greater access to $\pi^{\mathrm{cloud}}$ permits both shorter local reasoning and fewer environment steps while maintaining the reward constraint. At $C_D^{\max}=0.90$, a single operating point dominates the alternatives, so the frontier collapses to one point.

\subsection{Code Generation: MBPP}
\label{app:mbpp_details}

This section provides additional experimental and calibration details for the MBPP
evaluation in Sec.~\ref{sec:exp_mbpp}. Table~\ref{tab:mbpp_test} summarizes the
held-out performance of each policy across repeated calibration--test splits.

\begin{table}[t]
\centering
\caption{MBPP test-set results averaged over $50$ random $60/40$
calibration--test splits, with $n_{\mathrm{test}}=103$ problems per split.
For each split, $\hat{R}$ is the empirical test reward, $\hat{C}_D$ is the
fraction of test problems deferred to the cloud model, and $\hat{C}_L$ is the
mean number of edge-model thinking tokens per test problem. The bold row
denotes TSDS evaluated at the operating point selected by LTT on the
corresponding calibration partition. A dash indicates that the metric is not
applicable because Cloud ReAct performs no edge-model reasoning.}
\label{tab:mbpp_test}
\small
\begin{tabular}{lccc}
\toprule
Policy & $\hat{R}$ & $\hat{C}_D$ & $\hat{C}_L$ (tokens) \\
\midrule
E-ReAct
    & $0.719$ & $0.000$ & $1183$ \\
Cloud ReAct
    & $0.810$ & $1.000$ & \text{---} \\
E-ReAct-TC ($\lambda_L=0.80$)
    & $0.620$ & $0.000$ & $422$ \\
ReDAct-CD ($\lambda_L=\infty$)
    & $0.779$ & $0.786$ & $1206$ \\
\textbf{TSDS ($\hat{\boldsymbol{\lambda}}$)}
    & $\mathbf{0.714}$ & $\mathbf{0.346}$ & $\mathbf{422}$ \\
\bottomrule
\end{tabular}
\end{table}

\paragraph{Aggregate performance:}
TSDS attains the same mean thinking cost as E-ReAct-TC while increasing the
mean test reward from $0.620$ to $0.714$ through selective cloud deferral.
Relative to ReDAct-CD, TSDS reduces the mean number of edge-model thinking
tokens by approximately $65\%$ ($422$ versus $1206$) and satisfies the
cloud-deferral constraint, whereas ReDAct-CD exceeds the budget with
$\hat{C}_D=0.786$. These results illustrate the benefit of jointly calibrating
thought truncation and cloud deferral: truncation reduces local computation,
while selective deferral recovers much of the reward lost by truncation alone.

\paragraph{Repeated-split evaluation:}
To quantify sensitivity to the calibration--test partition, we sample
$N_{\mathrm{splits}}=50$ random $60/40$ partitions of the $257$ MBPP
evaluation problems. Each partition contains $154$ calibration problems and
$103$ held-out test problems. For every partition, we rerun the complete LTT
procedure using only the calibration portion, while keeping all other
experimental settings fixed, and evaluate the selected operating point on the
corresponding test portion. Reported means and standard deviations are
computed across the $50$ partitions. Under this protocol, TSDS achieves
a mean test reward of
$\hat{R}_{\mathrm{test}}=0.714\pm0.026$ and a mean thinking cost of
$\hat{C}_L=422\pm118$ tokens per problem.

\paragraph{Rollout and probe configuration:}
Both $\pi^{\mathrm{edge}}$ and $\pi^{\mathrm{cloud}}$ use a maximum reasoning
budget of \texttt{max\_thought}$=2048$ tokens. When allowed to complete its
reasoning trace, the 7B edge model generates an average of $1166$ thinking
tokens per problem. The 32B cloud model generally produces longer traces;
however, cloud-generated tokens are not included in $\hat{C}_L$, which
measures on-device reasoning cost. The EMA convergence probe is trained at
layer $\ell=5$ using $930$ position-level probe samples collected from $100$
MBPP training episodes. The PPL uncertainty score is computed from the
token-level probabilities of the generated Python function body.

\paragraph{Representative certification calculations:}
To make the LTT decisions explicit, we report the calibration results for the
representative seed-$0$ partition. This partition contains
$N_{\mathrm{cal}}=154$ problems, and the Bonferroni-corrected rejection
threshold is
\[
\frac{\delta}{|\Lambda|}
=
\frac{0.10}{45}
=
0.00222.
\]
The resulting decisions are as follows:
\begin{itemize}
    \item \emph{E-ReAct} solves $119/154$ calibration problems, giving
    $p^R=0.013>0.00222$; it therefore fails the corrected reward test.

    \item \emph{E-ReAct-TC}
    ($\lambda_L=0.80$, $\lambda_D=+\infty$) solves $111/154$ calibration
    problems, giving $p^R=0.21>0.00222$, and likewise fails the reward test.

    \item \emph{ReDAct-CD}
    ($\lambda_L=\infty$, $\lambda_D=0.084$) passes the reward test but has
    $\hat{C}_{D,\mathrm{cal}}=0.786>C_D^{\max}=0.70$, so it fails the
    cloud-deferral constraint.

    \item \emph{Cloud ReAct} defers on every calibration problem and therefore
    has $\hat{C}_{D,\mathrm{cal}}=1.000>C_D^{\max}$.

    \item \emph{TSDS}
    ($\hat{\boldsymbol{\lambda}}=(0.80,0.084)$) solves $123/154$ calibration
    problems, giving $p^R=0.0018<0.00222$. Its deferral-rate test also passes
    the corrected threshold, so the joint p-value
    $\max\{p^R,p^C\}$ is below $0.00222$ and the operating point receives
    joint LTT certification.
\end{itemize}
\subsection{Household Robot: Environment Layout and Diagnostics}
\label{app:household_details}

\paragraph{Motivation:} The custom text-world example adopted in Sec.~\ref{sec:exp_household} was motivated by the observation that the implemented DeepSeek-R1-Distill models, being trained primarily for mathematical and code-generation reasoning, achieve near-zero task completion on more complex benchmarks such as ALFWorld \citep{shridhar2021alfworld}.

\paragraph{Probe layer:}
\label{app:probe_layer_household}
On the household robot task, an ablation over $\ell \in \{5, 8, 10, 14, 20\}$ on
$\mathcal{D}_{\mathrm{tr}}$ shows the EMA probe reaches its best AUROC at $\ell=10$ ($0.863$)
versus $0.830$ at $\ell=5$.  This differs from the other three benchmarks where $\ell=5$ is
optimal, and confirms that the best probing layer is task-dependent rather than a fixed
architectural default.

\begin{figure}[t]
\centering
\includegraphics[width=0.55\textwidth]{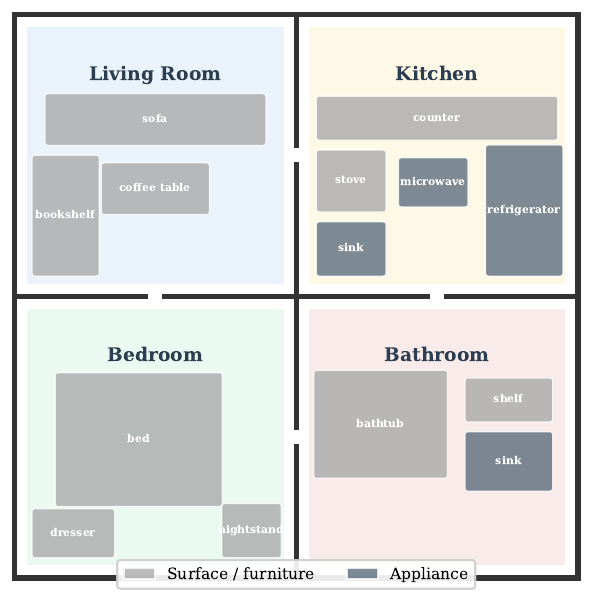}
\caption{Floor plan of the simulated household environment.
Each room contains a set of surfaces and appliances (see legend); objects are
placed on surfaces at episode initialization.
All four rooms are mutually accessible via the \texttt{go~to~<room>} action in a single
step, irrespective of the agent's current location.}
\label{fig:household_floorplan}
\end{figure}

Fig.~\ref{fig:household_floorplan} shows the floor plan of the simulated household
environment used in Section~\ref{sec:exp_household}.  The environment consists of four rooms
(living room, kitchen, bedroom, bathroom), each containing a fixed set of surfaces and
appliances on which objects are placed at the start of each episode.  The agent navigates
between rooms via \texttt{go~to~<room>} and interacts with objects via \texttt{pick~up},
\texttt{put~down}, \texttt{heat}, and \texttt{clean} actions, aiming to complete one of four
task families (fetch, place, clean, heat) within at most six steps.

\subsection{Comparison with Fixed Token-Budget Baselines}
\label{app:fixed_budget}

A natural question is whether TSDS's compute savings may also be obtained using probe-free static token caps \citep{muennighoff2025s1}.
We address this concern here by defining and evaluating a \emph{fixed token-budget} variant, \textbf{FT-CD} (Fixed-Token Calibrated Deferral), on the GSM8K data from Section~\ref{sec:experiments} (Appendix~\ref{app:gsm8k_details}).

\paragraph{Baseline definition.}
FT-CD is identical to TSDS except that the probe-driven stopping rule \eqref{eq:stopping} is replaced by deterministic truncation at a fixed position $p^{\mathrm{fix}}$,
independent of any hidden-state information.
The deferral rule, PPL uncertainty score, and multi-objective LTT procedure (Sec.~\ref{sec:multi_obj_calibration}) are unchanged.
We evaluate FT-CD on the same 1300-episode GSM8K rollout, the same 50 random $85/15$ calibration--test splits, and the same constraints ($R^{\min}{=}0.855$, $C_D^{\max}{=}0.70$, $\delta{=}0.10$, binomial $p$-values, Bonferroni correction with $|\Lambda^{\mathrm{FT}}|{=}20$ pairs, i.e.,\ $p^{\mathrm{fix}} \in \{48, 80, 112\}$ tokens against the same five $\lambda_D$ quantiles as TSDS).
FT-CD actions and PPL values at each fixed position are obtained from the pre-recorded strided early-stop outputs already present in the rollout files; no additional inference is required.

\paragraph{Results.}
Table~\ref{tab:ftcd} reports test-set reward, deferral rate, and mean thinking cost for FT-CD at each $p^{\mathrm{fix}}$ with $\lambda_D{=}0.088$ (the same threshold selected by TSDS), together with TSDS~$\hat{\boldsymbol{\lambda}}{=}(0.10,\,0.088)$.

\begin{table}[h]
\centering
\caption{GSM8K test-set comparison of TSDS and FT-CD at matched deferral threshold $\lambda_D{=}0.088$.
All methods use the same LTT procedure ($R^{\min}{=}0.855$, $C_D^{\max}{=}0.70$, $\delta{=}0.10$).
Budget shows the thinking-compute parameter (fixed token cutoff $p$ for FT-CD; probe threshold $\lambda_L$ for TSDS).
$\hat{R}_{\mathrm{test}}$ shows the fraction of test problems solved correctly.
$\hat{C}_{D,\mathrm{test}}$ shows the fraction of test problems deferred to the cloud model.
$\hat{C}_{L,\mathrm{test}}$ shows the mean thought tokens per problem on the test set.}
\label{tab:ftcd}
\small
\begin{tabular}{lccccc}
\toprule
Method & Budget & $\lambda_D$ & $\hat{R}_{\mathrm{test}}$ & $\hat{C}_{D,\mathrm{test}}$ & $\hat{C}_{L,\mathrm{test}}$ \\
\midrule
$\pi_{\mathrm{edge}}$ (no deferral) & $\infty$ & $\infty$ & 0.856 & 0.000 & 179.7 \\
FT-CD & $p{=}48$ tok & 0.088  & 0.913 & 0.190 & 48.0 \\
FT-CD & $p{=}80$ tok & 0.088 & 0.933 & 0.159 & 79.7 \\
FT-CD & $p{=}112$ tok & 0.088  & 0.913 & 0.144 & 109.2 \\
\midrule
TSDS & $\lambda_L{=}0.10$ & 0.088 & \textbf{0.913} & 0.221 & \textbf{17.4} \\
\bottomrule
\end{tabular}
\end{table}

First, TSDS matches the best reward achieved by FT-CD ($\hat{R}_{\mathrm{test}}{=}0.913$) while using $2.8\times$ fewer thought tokens ($\hat{C}_{L,\mathrm{test}}{=}17.4$ vs.\ $48.0$).
The probe's adaptive stopping, i.e., halting easy problems early while allowing harder ones more time, achieves the same reward at substantially lower mean compute than the fixed-budget can.
This per-problem adaptivity is directly reflected in the EMA probe's AUROC of $0.815$ (Table~\ref{tab:exp1_arch}), well above the position-only floor of $0.530$ that represents the maximum discriminative power available to any fixed-budget rule.

Second, reward under FT-CD is non-monotone in the stopping token position $p$; it rises from $0.913$ at $p{=}48$ to $0.933$ at $p{=}80$, then falls back to $0.913$ at $p{=}112$, showing that allocating more thinking tokens beyond a task-specific saturation point does not improve and can even hurt performance.
TSDS avoids this problem by stopping as soon as the probe signals convergence, rather than committing to a fixed budget.

\section{Thought Calibration--Deferral Interplay: Probe Signal and PPL Across the Thought Trace}
\label{app:interplay}

\begin{figure}[h]
\centering
\includegraphics[width=\textwidth]{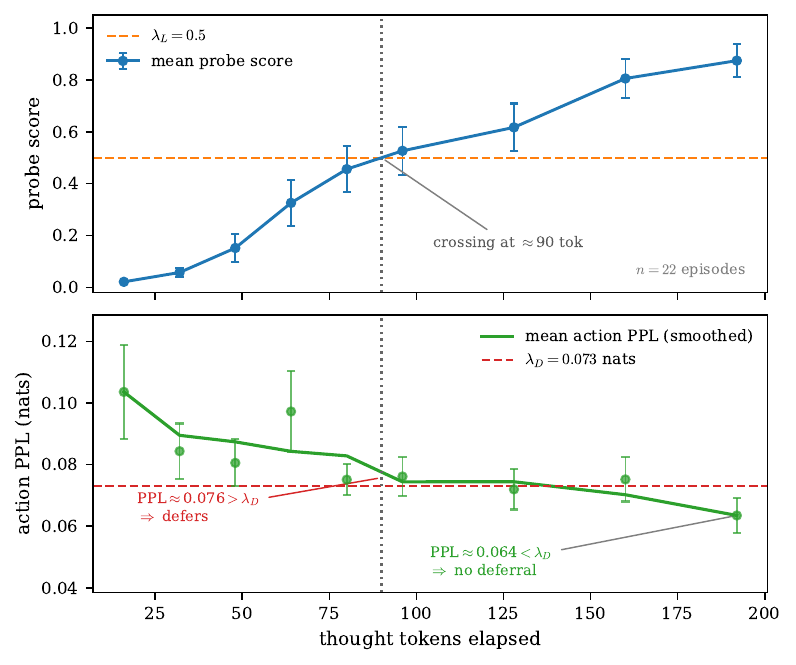}
\caption{Probe score (top) and action PPL (bottom) as a function of thought tokens elapsed,
averaged over GSM8K episodes in which the edge model's full-thought action is suboptimal, i.e., the final boxed answer does not match the gold answer.
The dashed orange and red lines mark the illustrative thresholds $\lambda_L$ and $\lambda_D$,
respectively; the dotted vertical line marks where the mean probe score crosses $\lambda_L$.
Error bars show ${\pm}1$ standard error.}
\label{fig:interplay}
\end{figure}

Fig.~\ref{fig:interplay} provides empirical evidence for the TC--deferral interplay described in
Section~\ref{sec:intro} and illustrated in Fig.~\ref{fig:framework}(c).
We focus on episodes in which the edge model's action is suboptimal. In the GSM8K context,
this means the final boxed integer produced after the complete thought trace does not match the
gold answer. These are the episodes where deferral is most needed.

The bottom panel shows that PPL decreases as reasoning continues. The edge model grows
progressively more confident in its (wrong) answer the longer it thinks.
At the point where the probe score crosses $\lambda_L$, where TSDS would halt reasoning , PPL
is still above $\lambda_D$, so deferral correctly fires and the episode is escalated to the cloud.
If reasoning instead runs to its full length, PPL drops below $\lambda_D$ and deferral no longer
triggers, allowing the suboptimal action to pass unchallenged.

This is the risk flagged in the introduction: extended thinking might not correct the error, and can suppresses the uncertainty signal that would otherwise catch it.
Jointly calibrating $(\lambda_L, \lambda_D)$ via the procedure in
Section~\ref{sec:multi_obj_calibration} ensures that TSDS halts at a point where the deferral
mechanism retains the ability to escalate uncertain, suboptimal actions to the cloud.
\end{document}